\documentclass[a4paper,fleqn]{cas-dc}

\usepackage[authoryear,longnamesfirst]{natbib}
\usepackage{caption}    
\usepackage{graphicx}
\usepackage{subcaption}
\usepackage{float}
\usepackage{tabularx} 
\usepackage{subcaption}
\def\tsc#1{\csdef{#1}{\textsc{\lowercase{#1}}\xspace}}
\tsc{WGM}
\tsc{QE}

\begin{document}
\let\WriteBookmarks\relax
\def\floatpagepagefraction{1}
\def\textpagefraction{.001}

\shorttitle{}    



\title [mode = title]{An Intelligent Decision Support System for Emotion Monitoring using Microscopic Fixational Dynamics}

 \fntext[1]{}
\author[1]{Xiangyu Shen}
\cormark[1] 
\ead{shenxiangyu@csuft.edu.cn}
\cortext[1]{Corresponding author} 

\author[2]{Feiyang Deng}

\author[1]{Zijian Dai}
\ead{20231200571@csuft.edu.cn}

\author[1]{Aibin Chen}
\ead{chenaibin@csuft.edu.cn}

\author[1]{Jizheng Yi}
\ead{T20152279@csuft.edu.cn}

\author[3]{Jie Li}
\ead{jieli.jsj@csuft.edu.cn}

\author[4]{Hongbo Jiang}
\ead{hongbojiang2004@gmail.com}

\affiliation[1]{organization={College of Advanced Interdisciplinary Studies, Central South University of Forestry and Technology},
            city={Changsha},
            postcode={410004}, 
            state={Hunan},
            country={China}}

\affiliation[2]{organization={Chow Yei Ching School of Graduate Studies, City University of Hong Kong},
            city={Hong Kong},
            postcode={999077}, 
            country={Hong Kong}}

\affiliation[3]{organization={College of Computer and Mathematics, Central South University of Forestry and Technology},
            city={Changsha},
            postcode={410004}, 
            state={Hunan},
            country={China}}

\affiliation[4]{organization={College of Computer Science and Electronic Engineering, Hunan University},
            city={Changsha},
            postcode={410082}, 
            state={Hunan},
            country={China}}

\begin{abstract}
	The rising prevalence of psychological disorders necessitates effective emotion monitoring, yet current methods relying on facial or physiological signals often suffer from intrusiveness and privacy issues. This paper proposes an intelligent decision support system and pervasive edge-computing framework that leverages smart glasses and a companion smartphone to infer emotional states from microscopic visual fixation patterns. Moving beyond traditional macroscopic gaze metrics, the proposed system extracts and decomposes three distinct neurophysiological micro-movements: microsaccades, ocular drifts, and ocular microtremors. We introduce an interpretable hybrid artificial intelligence pipeline---combining a multi-head attention mechanism, extreme gradient boosting, and a support vector machine---to extract deep temporal features, quantify their physiological importance, and perform efficient on-device classification. Through an extensive evaluation involving 60 volunteers, we rigorously validate the framework under a strict leave-one-subject-out cross-validation protocol across both controlled and naturalistic mobile scenarios. Ablation studies unequivocally demonstrate that these fixational micro-movements are substantially more discriminative for emotion inference than traditional macroscopic features. Furthermore, aligned with contemporary affective science, the system incorporates a few-shot personalization mechanism to bridge universal physiological baselines with individual emotional heterogeneity, achieving a highly robust personalized F1-score of 83.6\%. This work establishes a physiologically interpretable, unobtrusive, and deployable paradigm for continuous real-time emotion monitoring.
\end{abstract}






\maketitle

\section{Introduction}
\label{sec:introduction}
Emotion plays a fundamental role in human behavior and interaction, influencing various physiological responses, including heart rate, pupil dilation, and eye movement\citep{khare2024emotion}. Emotion recognition technology has become increasingly important for enhancing user experience, improving mental health monitoring, and enabling adaptive applications in human-computer interaction. However, traditional methods of emotion recognition, such as facial expression recognition and voice analysis, present challenges in privacy and practicality for real-world use. These methods often require direct observation of facial or vocal cues, which may be intrusive and limit user privacy\citep{canal2022survey,li2022eeg,zhang2024deep,liu2023multi}. 

In this context, we raise a question: \textit{can we capture users’ emotions through spontaneous physiological activities without violating their privacy?} Recent studies indicate that when humans are in a state of fixation, their eyes are not entirely stationary; instead, they exhibit three types of subtle eye movements---microsaccades, ocular drifts, and ocular microtremors---that are closely associated with autonomic emotional states~\citep{rucci2015control,alexander2019fixational,rucci2015fixational,yang2023natural}. In other words, these previously overlooked fixation patterns contain emotion-related physiological markers that can be extracted and analyzed. Simultaneously, the ubiquity of smart glasses is growing rapidly. According to the International Data Corporation (IDC), global smart glasses shipments in the first quarter of 2025 reached 1.487 million units, an 82.3\% year-on-year increase~\citep{IDC}. Therefore, we consider leveraging the built-in cameras of smart glasses to record users' microscopic fixation patterns during daily wear, enabling unobtrusive emotion monitoring. If successfully implemented, this fixation-pattern-based mobile paradigm offers unique advantages:

\begin{itemize}
	\item \textbf{Non-intrusive and privacy-preserving:} Unlike methods relying on facial expressions or voice recordings, analyzing subtle fixational micro-movements avoids capturing sensitive environmental or identifiable biometric data, providing a highly privacy-conscious approach.
	
	\item \textbf{Fine-grained, physiologically-grounded detection:} Macroscopic behaviors (e.g., facial expressions) can be consciously suppressed or faked. In contrast, microscopic eye movements are involuntary autonomic reflexes. Analyzing these patterns enables the detection of nuanced physiological shifts in the valence-arousal space~\citep{kollias2022abaw,yik2023relationship,martinez2025using}.
	
	\item \textbf{Real-world mobile applicability:} Utilizing commodity smart glasses allows for continuous emotion monitoring in diverse everyday contexts, from casual social interactions to immersive virtual experiences, without requiring highly specialized or intrusive medical-grade sensors.
\end{itemize}

Achieving this goal, however, entails significant technical and theoretical challenges. First, from a systems perspective, extracting millisecond-level micro-movements typically requires heavy computational resources, which contradicts the severe power and thermal constraints of wearable devices. Second, from an algorithmic perspective, deep learning models often act as "black boxes," making it difficult to interpret which specific physiological features actually drive the emotion classification. Finally, contemporary affective science---such as the \textit{Theory of Constructed Emotion}~\citep{barrett2017theory}---posits that emotions are not strictly universal but exhibit significant inter-individual variability, complicating the deployment of "one-size-fits-all" models in the wild.

To address these challenges, we propose \textbf{EmoGaze}, an intelligent edge-based expert system that pioneers the extraction and classification of microscopic fixation patterns for emotion monitoring. Unlike conventional pervasive sensors, EmoGaze is designed as an \textit{Intelligent Decision Support System (IDSS)} tailored for domains requiring unobtrusive affective profiling, such as clinical psychological assessment, driver vigilance monitoring, and adaptive human-computer interaction. To ensure deployability, EmoGaze partitions the workload: the smart glasses function purely as a lightweight sensor node, transmitting video streams to a companion computing hub. Algorithmically, we engineer an interpretable expert pipeline combining a Multi-Head Attention (MHA) mechanism, XGBoost, and a Support Vector Machine (SVM). This hybrid architecture not only achieves high accuracy but also acts as an "explainable AI (XAI)" module, quantitatively revealing the physiological importance of each micro-movement modality to domain experts.

We have prototyped EmoGaze and conducted extensive evaluations involving 60 volunteers across both controlled and naturalistic mobile scenarios. Evaluated under a rigorous LOSO-CV protocol, EmoGaze demonstrates robust zero-shot generalization to unseen individuals, with performance improving significantly to an F1-score of 83.6\% \footnote{Our study was approved by our university IRB. It did not raise any ethical issues.} upon applying minimal few-shot personalized calibration. In summary, our main contributions are as follows:

\begin{itemize}
	\item \textbf{Conceptual Innovation via Microscopic Signal Definition:} This research fundamentally reframes the input signal for emotion recognition. By decomposing fixation into its constituent neurophysiological components (microsaccades, ocular drifts, and microtremors), and validating their superiority over macroscopic gaze features through extensive modality ablation studies, we establish that this overlooked data stream contains robust, highly discriminative markers for complex emotional states.
	
	\item \textbf{Edge-Centric, Interpretable Hybrid Architecture:} We introduce a novel MHA-XGBoost-SVM pipeline meticulously designed for mobile edge computing. This architecture optimally balances deep non-linear representation power with deployment efficiency. Crucially, by utilizing XGBoost for feature importance ranking, we add a transparent layer of physiological interpretability to the "black box" attention features, empirically mapping specific micro-movements to high- and low-arousal states.
	
	\item \textbf{Explainable Expert System Paradigm:} Challenging the prevailing trend of "black-box" deep learning, EmoGaze demonstrates a highly interpretable expert system paradigm. We rigorously evaluate the system under a strict LOSO-CV protocol. By incorporating a few-shot personalization mechanism and explicitly extracting physiological features, our framework aligns with contemporary affective science, effectively bridging universal physiological baselines with individual emotional heterogeneity. This provides domain experts (e.g., clinicians, HCI designers) with transparent, data-driven insights into human affective states.
\end{itemize}

The rest of this paper is organized as follows. We review related work in Sec.~\ref{sec_related_work}. The physiological relationship between fixation patterns and emotions is explored in Sec.~\ref{sec_motivation}. We detail the EmoGaze system architecture in Sec.~\ref{sec_design}. Performance and system overhead evaluations are presented in Sec.~\ref{sec_evaluation}. Limitations and future directions are discussed in Sec.~\ref{sec_discussion}, and we conclude in Sec.~\ref{sec_conclusion}.

\section{Related Work}
\label{sec_related_work}

In this section, we review the efforts of researchers in mobile emotion detection and provide a comprehensive overview of EmoGaze's advantages compared to state-of-the-art works.

\subsection{Diverse Emotion Detection}
In recent years, emotion detection has gained considerable attention, particularly in contexts like human-computer interaction, mental health monitoring, and user experience optimization \citep{acheampong2021transformer,mao2022biases,nie2023long}. Researchers have explored various approaches to detect emotions, utilizing modalities such as facial expressions, speech, physiological signals, eye movements, and multimodal data \citep{deng2020multi,awais2020lstm,zhang2020emotion,chen2021emoji,arun2023facial,kulkarni2025hybrid}. 

Facial expression recognition remains one of the most widely used methods, leveraging computer vision techniques such as convolutional neural networks (CNNs) \citep{zhang2018facial,li2020attention}. However, facial expression-based methods face severe limitations when dealing with face occlusion, varied lighting conditions, or individuals intentionally suppressing their expressions~\citep{zhang2018facial}. Speech-based emotion recognition utilizes features like tone and pitch \citep{khalil2019speech}. Yet, it is continuously challenged by environmental background noise, dialect variations, and the lack of continuous verbal communication in daily life. Physiological signal-based methods (e.g., ECG, EMG, EEG) offer high precision by directly measuring bodily responses \citep{song2020emotion}. However, they are inherently intrusive, requiring specialized medical-grade equipment that severely limits their deployment in pervasive mobile applications \citep{yin2017recognition}.

Eye-tracking data has emerged as a highly promising modality, providing unobtrusive insights into emotional states. Recent works have demonstrated that macroscopic eye movement features, such as fixation duration, saccade amplitude, and pupil dilation, are sensitive indicators of emotional responses \citep{abdou2022gaze}. However, macroscopic gaze metrics are often heavily confounded by external visual tasks, ambient lighting (affecting the pupil), and cognitive load, which complicates reliable emotion detection in unconstrained real-world settings~\citep{wu2021emotion}. Furthermore, multimodal approaches \citep{zhang2024deep,lv2021progressive,middya2022deep} attempt to fuse these diverse signals to improve robustness. Unfortunately, collecting and synchronously processing data from multiple high-bandwidth sensors (e.g., cameras and EEG) poses prohibitive computational overhead for mobile and edge computing platforms.

In contrast to the aforementioned methods, this work introduces a novel approach based on the fine-grained microscopic patterns of visual fixations. Crucially, while macroscopic expressions (e.g., facial muscle movements or large saccades) can be consciously controlled or easily disrupted by task demands, the microscopic dynamics within a single fixation are involuntary autonomic reflexes. By extracting these subtle, task-independent physiological markers, we aim to establish a robust and computationally lightweight framework for continuous, on-the-go emotion monitoring.

\begin{table*}[h]
	\caption{Gaze-based Emotion Recognition Methods}
	\label{gaze_based_emotion_recognition}
	\centering
	\resizebox{\textwidth}{!}{%
		\begin{tabular}{llll}
			\toprule
			Related Works                         & Method                                       & Data Modality                                                               & Limitations/Advantages                                                                                                                                   \\ \midrule
			Ashwaq Alhargan et al.~\citep{alhargan2017multimodal} & Support Vector Machine + multimodal fusion                    & \begin{tabular}[c]{@{}l@{}}Eye tracking\\ Speech\end{tabular}               & Needs speech; shallow features                                                                                                                           \\
			\multicolumn{4}{l}{}                                                                                                                                                                                                                                                                                                  \\
			Bere~\citep{zhu2024bere}             & \begin{tabular}[c]{@{}l@{}}Graph Convolutional Network\\ Cross-Modal Domain Adaptation\end{tabular}                 & \begin{tabular}[c]{@{}l@{}}Eye tracking\\ Head movement\end{tabular}        & Non-emotion focus; lacks fine-grained eye modeling                                                                                                                  \\
			\multicolumn{4}{l}{}                                                                                                                                                                                                                                                                                                  \\
			VREED~\citep{tabbaa2021vreed}                      & Statistics + traditional machine learning                  & \begin{tabular}[c]{@{}l@{}}Eye tracking\\ ECG\\ GSR\end{tabular}            & Multimodal required; severe wearable hardware burden                                                                                                                     \\
			\multicolumn{4}{l}{}                                                                                                                                                                                                                                                                                                  \\
			Vehlen et al.~\citep{vehlen2023reduced}             & Clinical study and oxytocin intervention           & Eye tracking                                                                & No algorithmic modeling; clinical environment only                                                                                                                               \\
			\multicolumn{4}{l}{}                                                                                                                                                                                                                                                                                                  \\
			EyeSyn~\citep{lan2022eyesyn}                  & Psychology-driven synthetic modeling         & Synthetic eye tracking                                                      & Synthetic data only; lacks real-world evaluation                                                                                                                   \\
			\multicolumn{4}{l}{}                                                                                                                                                                                                                                                                                                  \\
			Wang Kay Ngai et al.~\citep{ngai2022emotion}         & CNN with multimodal fusion                   & \begin{tabular}[c]{@{}l@{}}EEG\\ Facial images\\ Eye tracking\end{tabular} & Heavy model; requires multiple intrusive sensors                                                                                                                   \\
			\multicolumn{4}{l}{}                                                                                                                                                                                                                                                                                                  \\
			\textbf{EmoGaze (Ours)}                 & Positional encoding + multi-head attention & Fixation \newline
			\begin{math}
				\left\{
				\begin{array}{l}
					\text{MS} \\
					\text{OD} \\
					\text{OMT}
				\end{array}
				\right.
			\end{math}                                                   & \begin{tabular}[c]{@{}l@{}}Lightweight and deployable on mobile edge devices;\\ Single-modality (dual-eye cameras) non-intrusive setup;\\ Fine-grained global fixation modeling via attention\end{tabular} \\ \bottomrule
		\end{tabular}%
	}
\end{table*}

\subsection{Connecting Fixation with Emotion and Contemporary Affective Science}
In recent years, the relationship between eye movement patterns and emotional states has gained considerable attention. While several studies have explored gaze patterns \citep{hadders2022human,cuve2021alexithymia,nag2020toward,black2020complex,le2020oxytocin}, most focus on macroscopic, behavioral gaze trajectories. A key contribution in this domain is the work by Yitzhak et al.~\citep{yitzhak2022many}, demonstrating that emotional states modulate overall visual attention and saccadic patterns. 

However, mapping complex emotions directly from macroscopic behaviors is being increasingly challenged by contemporary affective science. Notably, the \textit{Theory of Constructed Emotion} \citep{barrett2017theory} argues that emotions are not universally fixed, rigid templates (e.g., a specific facial expression always meaning "anger"). Instead, emotions are highly individualized, context-dependent phenomena constructed from deeper, low-level physiological states of arousal and valence.

Addressing this paradigm shift, our approach deliberately moves away from macroscopic gaze behavior. Instead, we hypothesize that the eye undergoes subtle yet measurable involuntary movements---namely microsaccades, ocular drifts, and ocular microtremors---that serve as the fundamental physiological substrate for arousal and valence. Microsaccades, small involuntary jerks, have been linked to cognitive load and autonomic arousal shifts~\citep{liu2022functional,yu2022microsaccades,liu2023microsaccades}. Ocular drifts have been shown to correlate with perceptual processing and attentional disengagement~\citep{malevich2020rapid,khademi2024visual}. Ocular microtremors are believed to be a direct byproduct of the central nervous system's regulation~\citep{graham2023ocular,leigh2013tremor,bolger1999ocular}. 

By decomposing and analyzing these three fundamental micro-movements, EmoGaze captures the purest, low-level physiological indicators of emotion. This aligns seamlessly with modern affective theories: rather than assuming a "one-size-fits-all" mapping, we extract robust underlying physiological markers that, when coupled with brief personalized calibration, can accurately capture the user's uniquely constructed emotional state.

\subsection{Gaze-based Emotion Recognition}	
To provide a clear overview of EmoGaze in relation to recent advancements in gaze-based emotion recognition, Table~\ref{gaze_based_emotion_recognition} presents a comparative summary. It highlights key aspects such as core methodologies, specific data modalities, and architectural limitations. Due to considerable heterogeneity in experimental designs (e.g., many studies lacking strict subject-independent cross-validation), direct comparison of accuracy figures is omitted. Instead, the primary aim of this table is to delineate the unique contributions of EmoGaze: it is a pioneering single-modality system that extracts deep temporal features from previously overlooked microscopic fixational dynamics, specifically engineered for deployment on mobile edge-computing platforms (e.g., smart glasses and companion smartphones).

\section{Motivation}
\label{sec_motivation}
\begin{figure*}[t]
	\hspace{0cm}  
	\centerline{\includegraphics[width=0.9\textwidth]{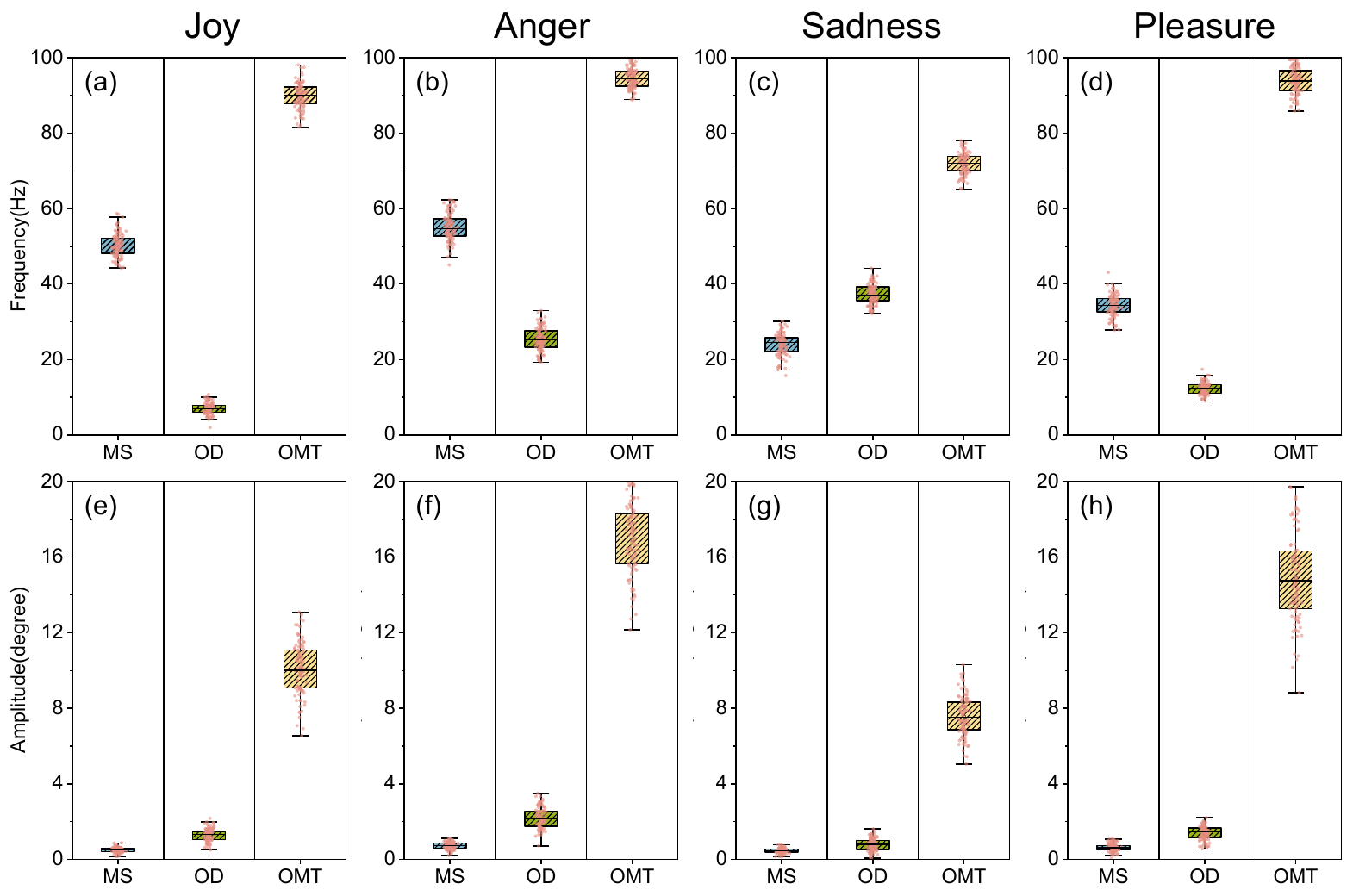}}
	\hspace{0cm}  
	\caption{Amplitude (degree) and frequency (Hz) of microsaccade (MS), ocular drift (OD), and ocular microtremor (OMT) under four basic emotions of joy, anger, sadness and pleasure. The frequency and amplitude of eye movement signals vary with different emotional states. High-frequency and high-amplitude signals are observed during anger, indicating heightened activity, while sadness shows lower frequency and amplitude, reflecting reduced activity. Joy and pleasure states have moderate signal levels, showing stable eye movements. The variability in eye movement signals is higher in anger, joy, and pleasure, but more stable in sadness.}
	\label{fig:motivation}
\end{figure*}

In this section, we explore the intricate relationship between eye fixation patterns and emotional states from physiological and neuroscientific perspectives, conducting a qualitative pilot study to establish this connection. 

Emotional cognition research posits that ocular motion metrics---such as microsaccades, ocular drifts, and ocular microtremors---offer significant insights into the underlying states of autonomic arousal and valence \citep{kashihara2014emotional,krejtz2020pupillary,lin2023cognitive,bolger2001effect,simola2015affective}. For instance, microsaccades, which are rapid, minute eye movements, exhibit increased frequency and amplitude during states of heightened emotional arousal, typically associated with anger and joy \citep{kashihara2014emotional,chen2021investigating,krejtz2020pupillary}. This increase reflects the autonomic nervous system's response to intense stimuli, serving as physiological correlates that mirror the central nervous system's activation. Conversely, ocular drifts---slow, involuntary eye movements---are observed to be more pronounced during emotional states characterized by low arousal and valence, such as sadness \citep{aston2005integrative,russell1980circumplex,nolen2008rethinking}. This phenomenon may be attributed to a reduced capacity for visual fixation control in depressed emotional states, reflecting a visual disengagement from external stimuli and a decrease in attentional focus.

Ocular microtremors, though less extensively studied compared to other ocular metrics, represent continuous, extremely small, high-frequency oscillations of the eyes. These involuntary movements are believed to reflect underlying cortical activity, suggesting a potential association with emotional fluctuations. Given their link to central nervous system processes, ocular microtremors provide valuable insight into the subtle physiological dynamics accompanying shifts in emotional states. Preliminary evidence indicates that changes in the frequency and amplitude of these tremors can be indicative of shifts in arousal and attentional focus \citep{graham2023ocular,robertson2007non,bolger2001effect,ryle2009compact}.

While contemporary affective science, such as the \textit{Theory of Constructed Emotion}~\citep{barrett2017theory}, emphasizes that final emotional experiences are context-dependent and exhibit inter-individual variability, these autonomic ocular micro-movements provide a universally grounded physiological substrate. By systematically mining these intricate physiological behaviors, we can capture high-fidelity markers of arousal and valence, setting a robust baseline that can subsequently accommodate personalized emotion construction.

To empirically illustrate how these fixational micro-patterns correlate with emotional states, we recruited an independent cohort of 14 volunteers (7 males and 7 females, aged from 20 to 53; completely distinct from the 60 subjects used in the main system evaluation to strictly prevent data leakage) to measure their fixation patterns while experiencing discrete emotions.

Specifically, we observe that in the anger state, the frequency and amplitude of all three eye movement signals are relatively high, reflecting the heightened physiological activity during extreme arousal. In contrast, during the sadness state, the frequency and amplitude of the signals are significantly lower, indicating a reduction in eye movement stability during emotional downturns. In the joy and pleasure states, the signals maintain moderate levels, representing stable ocular patterns under positive conditions. Moreover, the width of the box plots (i.e., variance) demonstrates that signal variability is higher in high-arousal states (anger, joy, pleasure), whereas sadness exhibits a more suppressed and concentrated eye movement behavior. 

This analysis confirms a definitive correlation between these three micro-eye movement signals and underlying emotional states, validating our premise that microscopic fixational dynamics contain rich, discriminative emotion indicators. Therefore, it is theoretically feasible and scientifically rigorous to infer users' emotions through deep algorithmic mining of their fixation patterns.

In the following section, we elaborate on the design of EmoGaze, an edge-centric hybrid deep-learning framework tailored to extract and classify these subtle fixation micro-movements.

\section{System Design}
\label{sec_design}

\subsection{Design Overview}
\label{subsec_overveiw}

\begin{figure*}[h]
	\vspace{-1ex}
	\centering
	\centerline{\includegraphics[width=1\textwidth]{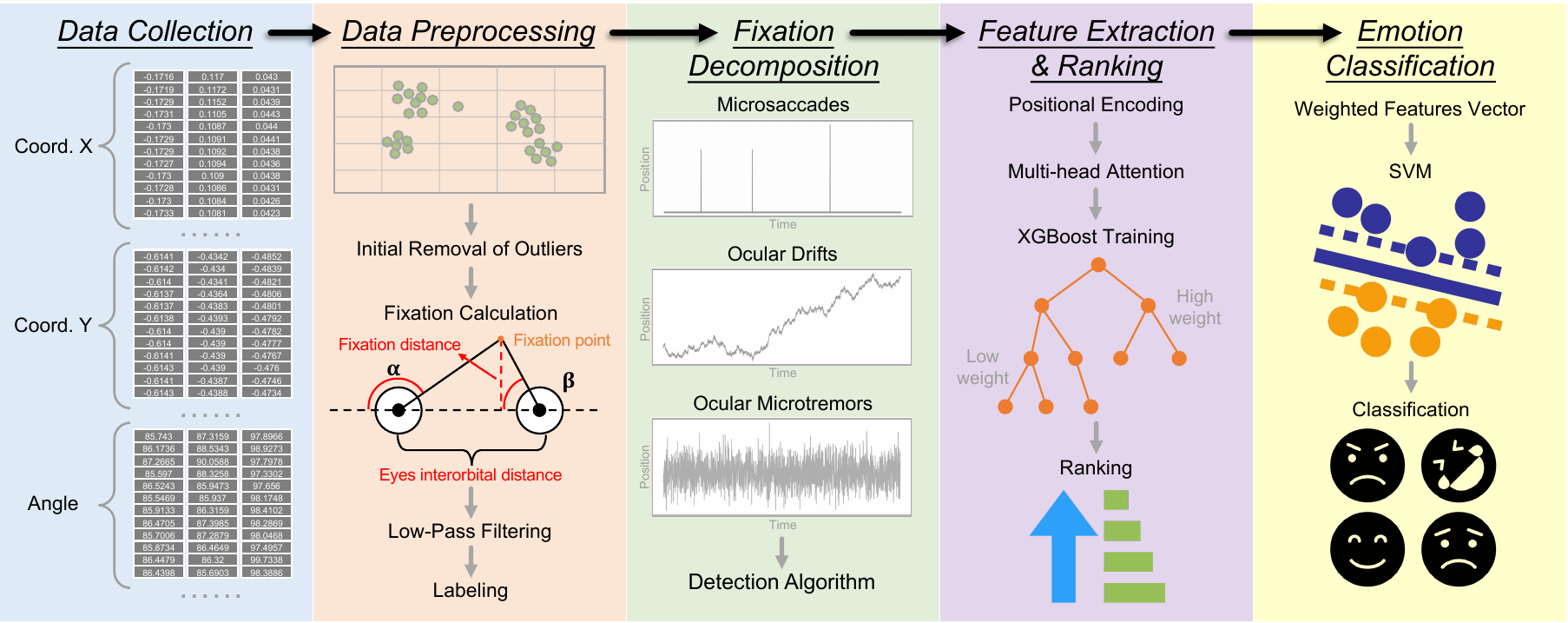}}
\caption{\textbf{EmoGaze involves five modules.} \textbf{Data Collection}: dual-eye ROI infrared video streams are captured by our custom-designed eye-tracking glasses and transmitted to the edge device. \textbf{Data Preprocessing}: X, Y coordinates are extracted via DSP, and raw fixation data are filtered according to physiological characteristics. \textbf{Fixation Decomposition}: the preprocessed fixation sequences are decomposed into three subtle eye movement signals by our detection algorithm. \textbf{Feature Extraction \& Ranking}: features are extracted through positional encoding and multi-head attention mechanisms, and weighted by XGBoost. \textbf{Emotion Classification}: emotions are classified by an SVM based on the weighted features.}
	\label{fig:system_overview}
\end{figure*}

The system overview of EmoGaze is illustrated in Fig.~\ref{fig:system_overview}. To align with the power and thermal constraints of wearable technology, EmoGaze adopts an edge-computing architecture that partitions tasks between a sensor node and a computational hub. The framework consists of five main modules: \emph{Data Collection}, \emph{Data Preprocessing}, \emph{Fixation Decomposition}, \emph{Feature Extraction \& Ranking}, and \emph{Emotion Classification}. 

In \emph{Data Collection}, the raw ocular data are acquired by our custom-designed smart glasses (Fig.~\ref{fig:eye_tracking_equipment}). Functioning purely as a high-fidelity sensor node, the glasses capture cropped Region of Interest (ROI) infrared video streams of both eyes. These video streams are transmitted via high-bandwidth Wi-Fi Direct (802.11ax) to a companion smartphone acting as the edge hub. In \emph{Data Preprocessing}, the smartphone first utilizes its Digital Signal Processor (DSP) to extract the X, Y coordinates and rotational angles of the pupils using a sub-pixel tracking algorithm. The system then filters outliers, removes unreasonable fixation distances, applies low-pass filtering, and labels the data. Building on this, in \emph{Fixation Decomposition}, specific algorithmic strategies extract the three subtle eye movement signals: ocular drifts, ocular microtremors, and microsaccades. In \emph{Feature Extraction \& Ranking}, positional encoding and a multi-head attention mechanism extract deep temporal features from these sequences. An XGBoost model is then utilized to evaluate feature importance and assign corresponding weights. Finally, in \emph{Emotion Classification}, a Support Vector Machine (SVM) utilizes these weighted features to infer emotional states. We elaborate on the algorithmic modules executed on the edge device in the following sections.

\subsection{Data Preprocessing}
\label{subsec_data_preprocessing}
After obtaining the fixation data, EmoGaze further performs data preprocessing, which includes removing outliers of left and right pupil angles, eliminating points with unreasonable fixation distances, applying low-pass filtering, and labeling the data. 

\subsubsection{Initial Removal of Outliers}
Since our device can obtain the rotational angle data of both left and right eyes (as shown by \(\alpha\) and \(\beta\) in Fig.~\ref{fig:system_overview}), we can remove abnormal rotational angle values. As illustrated in Fig.~\ref{fig:system_overview}, based on the triangle's exterior angle being equal to the sum of the non-adjacent interior angles, \(\alpha\) must be greater than \(\beta\). Therefore, we can remove fixation point data where \(\alpha \leq \beta\) to initially eliminate outliers.

\subsubsection{Fixation Distance Filtering }
As shown in Fig.~\ref{fig:system_overview}, we calculate the distance \(D\) from the fixation point to the eyes using the parallax method commonly used in astronomy\citep{yu2024smooth}. Assuming the interocular distance \(p\) is known, based on the properties of triangles, we can obtain \(D = \frac{p \cdot \tan(\alpha) \cdot \tan(\beta)}{\tan(\alpha) + \tan(\beta)}\). Therefore, we remove points with unreasonable fixation distances (those exceeding three times the standard deviation). This is because the human fixation point cannot undergo very large sudden changes.

\subsubsection{Low-Pass Filtering}
Since the frequency of eye movement signals is generally below 104Hz\citep{barea2002system}, we set a low-pass filter to reset the frequency coefficients above 104Hz to 0, thereby filtering out high-frequency noise. The signal is then converted from the frequency domain back to the time domain.

\subsubsection{Labeling and Expert Ground Truth Protocol}
To ensure the clinical and psychological validity of the emotion labels, participants completed the Positive and Negative Affect Schedule (PANAS)~\citep{watson1988development} alongside the Self-Assessment Manikin (SAM) ~\citep{bradley1994measuring} arousal scale immediately following each stimulus trial. PANAS provides two independent scores (Positive Affect, PA, and Negative Affect, NA), while SAM quantifies the physiological arousal level.

To map these continuous scores to our four discrete emotion categories, we utilize a mapping protocol informed by the Circumplex Model of Affect ~\citep{russell1980circumplex}. Using a subject-specific median split to establish thresholds (for practical deployment involving completely new users, a global population median is utilized during the initial cold-start phase before transitioning to individualized thresholds during few-shot calibration), the emotions are categorized as follows: \textbf{Joy} is characterized by high PA and high arousal; \textbf{Pleasure} represents high PA but low/moderate arousal; \textbf{Anger} maps to high NA and high arousal; and \textbf{Sadness} corresponds to high NA and low arousal. To eliminate ambiguity, participants also provided a forced-choice categorical label (Joy, Pleasure, Anger, Sadness). Samples where the categorical label contradicted the PANAS/SAM dimensional mapping are marked as ambiguous and discarded, yielding a highly reliable, expert-grade ground truth dataset for model training.


\subsection{Fixation Decomposition and Sensing Fidelity}
In Sec.~\ref{sec_motivation}, we introduce the three types of eye movement signals associated with fixation: microsaccades, ocular drifts, and ocular microtremors. In this section, we detail the separation of these signals from the raw fixation sequence and rigorously validate the sensing fidelity of our decomposition pipeline.

\subsubsection{Sensing Fidelity and Noise Floor Calibration}
A critical challenge in tracking fine-grained micro-eye movements on mobile wearables is distinguishing genuine physiological signals from hardware noise, especially given the minute amplitudes of Ocular Microtremors (OMT). While our cameras operate at a nominal resolution of 360~\!p to conserve power and bandwidth, we employ a widely-validated \textbf{sub-pixel center-of-mass pupil tracking algorithm}~\citep{kassner2014pupil,swirski2012robust} that enhances the effective angular resolution to less than $0.05^\circ$.

Furthermore, to unequivocally prove that the extracted high-frequency components are physiological rather than instrumental artifacts, we conduct a baseline noise floor calibration using a stationary artificial eye model under identical lighting conditions. Spectral analysis revealed that the systemic hardware and algorithmic jitter produced a baseline noise floor Power Spectral Density (PSD) of $< 10^{-4} \text{ deg}^2/\text{Hz}$. In contrast, analysis of our in-vivo data confirmed that the extracted OMT (40--100Hz) exhibited peak PSDs in the range of $10^{-2} \text{ deg}^2/\text{Hz}$. This yields a Signal-to-Noise Ratio (SNR) exceeding 20~dB, empirically confirming that the kinematic amplitudes of EmoGaze's extractions capture genuine autonomic ocular reflexes rather than noisy proxies.

\subsubsection{Ocular Drifts and Microtremors Extraction}
Having verified the signal fidelity, we first extract the ocular drifts (OD) and ocular microtremors (OMT). Ocular drifts are characterized by slow, smooth motions generally occurring within the 0--40~\!Hz frequency range, whereas ocular microtremors are high-frequency, low-amplitude oscillations typically falling within the 40--100~\!Hz range \citep{ahissar2016possible}. Because our 500~\!fps camera yields a Nyquist frequency of 250~\!Hz, it safely encompasses these physiological bands. We apply dedicated, high-order Butterworth bandpass filters to the preprocessed positional signals to isolate these respective components without phase distortion.

\subsubsection{Microsaccade Extraction}
Microsaccades are rapid, corrective eye movements that maintain visual clarity \citep{martinez2009microsaccades}. We extract them based on precise velocity and acceleration thresholding. This process is fundamentally inspired by established microsaccade detection baselines (e.g., the robust Engbert \& Kliegl algorithm principles \citep{engbert2003microsaccades}) and is optimized for our mobile pipeline:

\begin{figure*}[t]
	\centering{\includegraphics[width=1\textwidth]{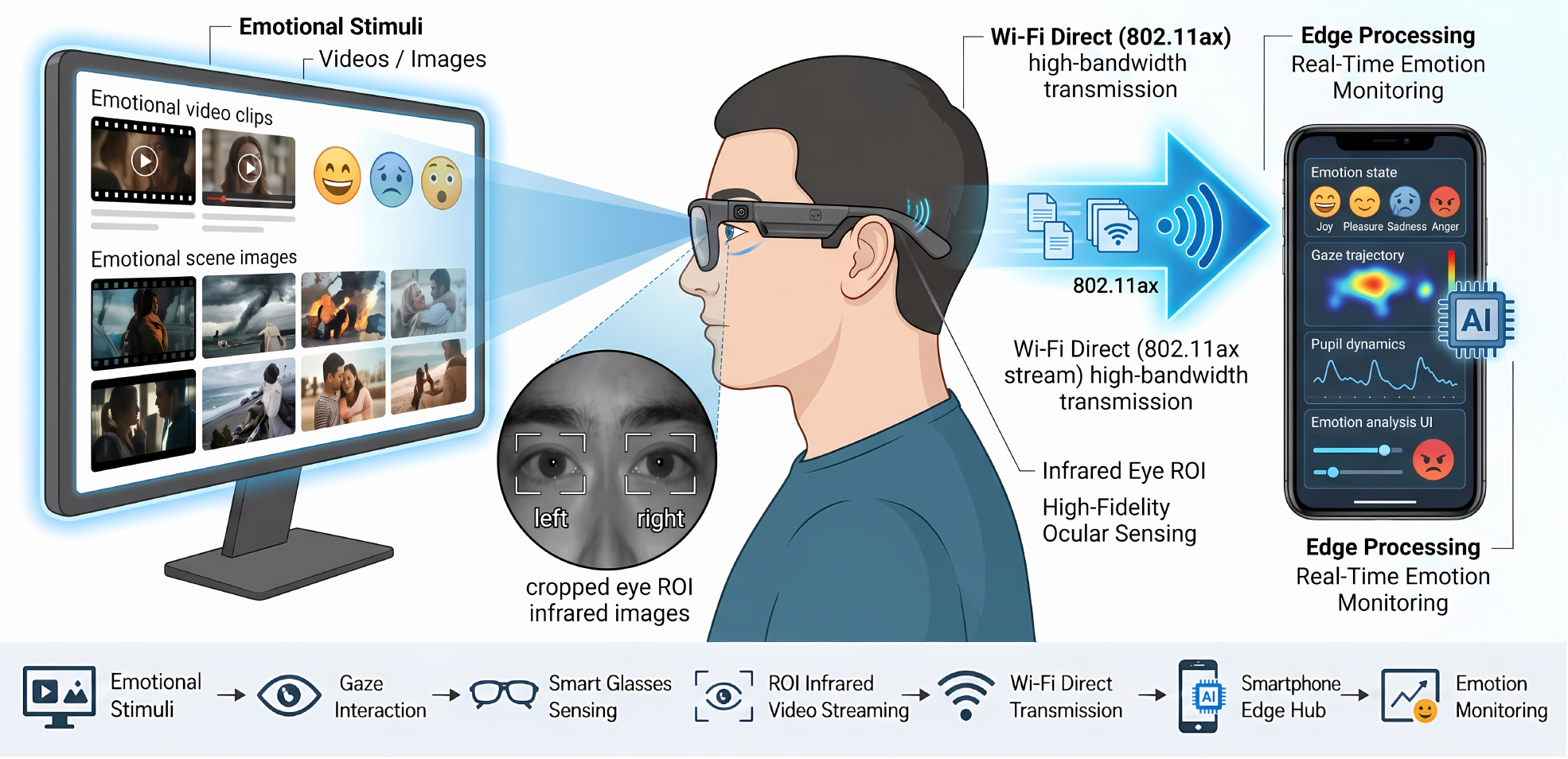}}
	\caption{Eye tracking equipment and corresponding software.}
	\label{fig:eye_tracking_equipment}
\end{figure*}

\textbf{1) Velocity Calculation:} 
Eye movement velocity is obtained by differentiating the position signals. The instantaneous velocity $v_t$ is calculated as Eq.~\ref{instantaneous_velocity} shows: 
\begin{equation}
	v_t=\frac{\sqrt{\left(x_t-x_{t-1}\right)^2+\left(y_t-y_{t-1}\right)^2}}{\Delta t},
	\label{instantaneous_velocity}
\end{equation}
where $x_t$, $y_t$ and $x_{t-1}$, $y_{t-1}$ represent the fixation coordinates at the current moment $t$ and preceding moment $t-1$. $\Delta t$ is the reciprocal of the sampling rate. The velocity is mapped to angular degrees per second ($^\circ/s$). This methodology adheres to the widely adopted velocity threshold (I-VT) principles \citep{olsen2012identifying}.

\textbf{2) Velocity Filtering:} 
To further isolate genuine microsaccades from transient high-frequency noise (e.g., blink artifacts), a sliding-window mean filter is applied to smooth the raw velocity data, defined in Eq.~\ref{filter}:
\begin{equation}
	v_{\text {filterd }}=\frac{1}{N} \sum_{i=0}^{N-1} v_{t-i},
	\label{filter}
\end{equation}
where $N$ denotes the sliding window sample size \citep{suthaharan2023microsaccade}.

\textbf{3) Event Screening and Detection:} 
We utilize a peak-finding algorithm to detect local velocity maxima. Because genuine physiological microsaccades exhibit very specific and short temporal durations (typically 40--70~\!ms \citep{taylor2024saccadic}), detected peaks failing to meet this strict temporal duration constraint are discarded.

\begin{figure*}[h]
	\centering{\includegraphics[width=1\textwidth]{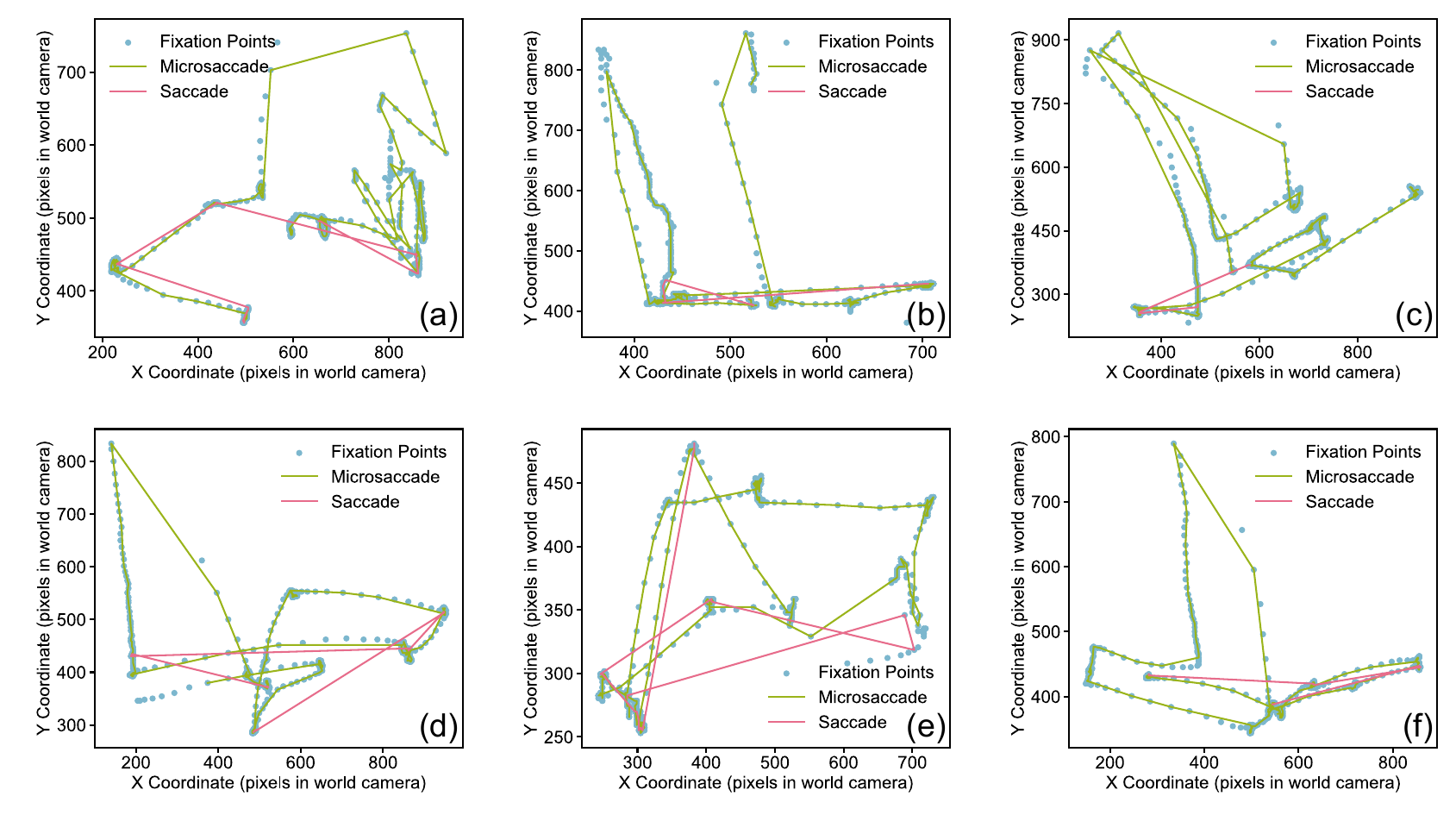}}
	\caption{Enhanced illustration of microsaccade detection, showcasing the effectiveness of our designed algorithm in distinguishing between microsaccades and saccades across multiple subfigures (a-f).}
	\label{fig:microssacade_detection}
\end{figure*}

\textbf{4) Event Verification:} 
To strictly differentiate microsaccades from deliberate macroscopic saccades, we evaluate instantaneous acceleration ($a_t$), computed via second-order differentiation (Eq.~\ref{accelaration}):
\begin{equation}
	a_t=\frac{v_t-v_{t-1}}{\Delta t}.
	\label{accelaration}
\end{equation}
Microsaccades exhibit smoother acceleration profiles, whereas macroscopic saccades produce abrupt spikes. Applying a slope threshold effectively separates these distinct mechanisms \citep{moller2002binocular}. As an external validation of our pipeline, the amplitude and duration distributions of the extracted MS perfectly align with established physiological norms (e.g., amplitudes $<1^\circ$). Fig.~\ref{fig:microssacade_detection} illustrates the successful decomposition of these signals.

\subsection{Feature Extraction \& Ranking}
\label{subsec_feature_extraction}
\subsubsection{Feature Extraction}
Although the coarse relationship between the three eye movement signals and emotional states has been mentioned in Sec.~\ref{sec_motivation}, it is challenging to directly determine emotional states through the sequence of these three eye movement signals. Therefore, we intend to explore the use of feature extraction techniques to identify features related to emotional states.

Since the three eye movement sequences obtained earlier are essentially time series, and the multi-head attention mechanism is well-suited for processing time series, the multi-head attention mechanism is proposed to extract emotion-related features from the three eye movement sequences.

\begin{figure*}[h]
	\centering{\includegraphics[width=1\textwidth]{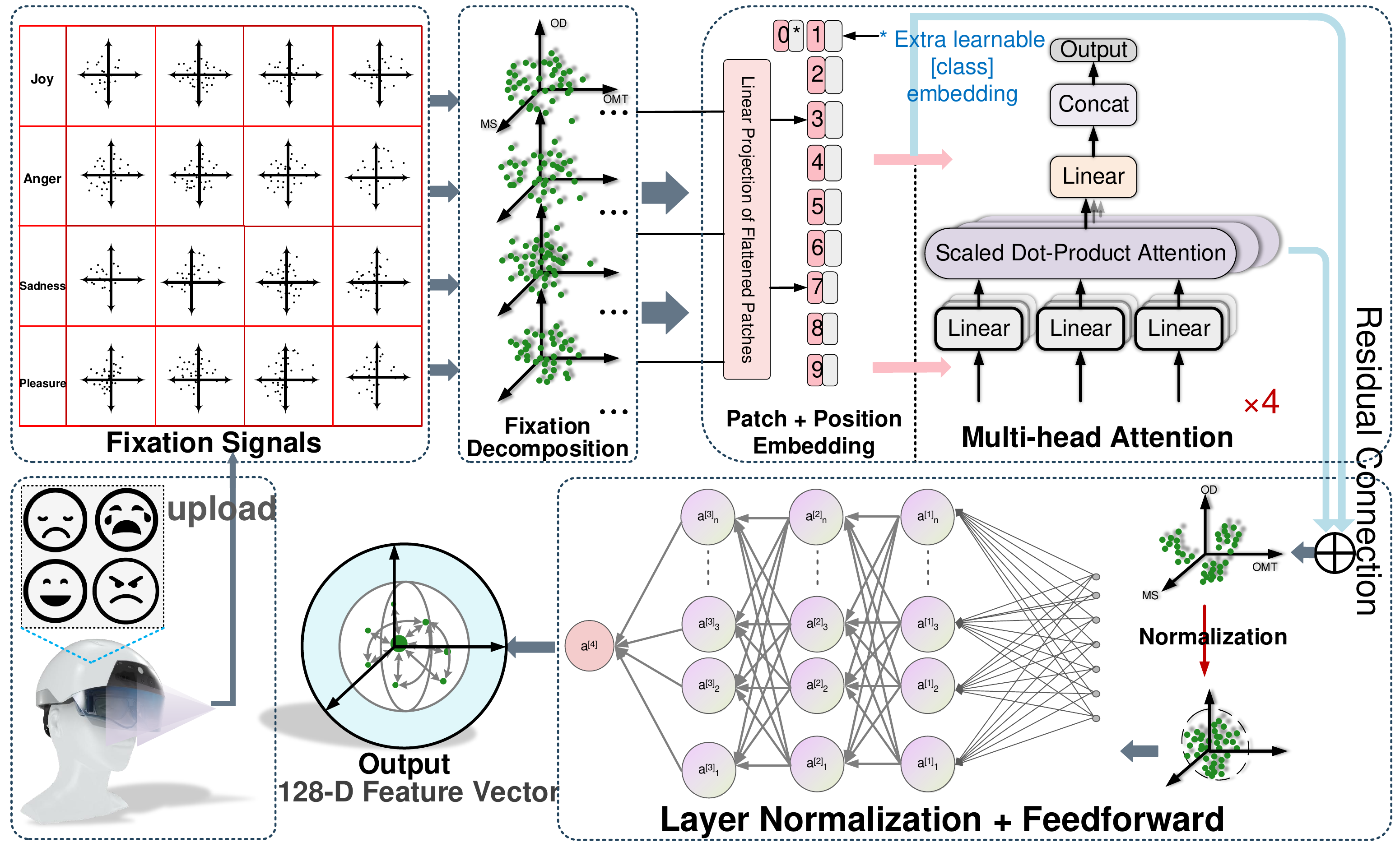}}
\caption{Detailed illustration of the feature extraction process based on multi-head attention. The framework maps input data to a higher-dimensional space, adds positional encoding, and processes the sequence through two stacked Transformer encoder layers. Each layer utilizes multi-head attention (4 attention heads, denoted by $\times 4$ in the figure) to capture global dependencies. After stabilizing training with residual connections and layer normalization, the features are aggregated into a global representation.}
	\label{fig:feature_extraction}
\end{figure*}

As shown in Fig.~\ref{fig:feature_extraction}, we propose a customized feature extraction framework that mining the three decomposed physiologically meaningful micro-movement components: Ocular Drift (OD), Microsaccades (MS), and Ocular Microtremor (OMT). This decomposition enables fine-grained modeling of distinct motion patterns within fixation signals.

Specifically, given an input fixation sequence of length $n$, we first decompose it into three independent feature matrices:
$F_{O D} \in \mathbb{R}^{n \times d_{OD}}$, $F_{M S} \in \mathbb{R}^{n \times d_{MS}}$, $F_{OMT} \in \mathbb{R}^{n \times d_{O M T}}$, where $d_{OD}$, $d_{MS}$ and $d_{OMT}$ represent the original feature dimension of OD, MS and OMT components respectively. To fully utilize the complementary information among these components, we perform a concatenation operation along the feature dimension:
\begin{equation}
	F=\operatorname{Concat}\left(F_{OD}, F_{MS}, F_{OMT}\right).
\end{equation}
The concatenated features are then projected into a unified embedding space through a linear transformation:
\begin{equation}
	X=F W_e+b_e,
\end{equation}
where $W_e\in \mathbb{R}^{\left(d_{OD}+d_{MS}+d_{OMT}\right) \times d}$ maps the combined features to the embedding dimension $d$. To preserve the temporal characteristics of each component, we inject sine position encoding $P$: $X^{\prime}=X+P, P \in \mathbb{R}^{n \times d}$, where $X^{\prime}$ represents the sequence of positional encoded component. 

Considering the differences in the physical properties of the different micromovement components, we have designed a dedicated multi-head attention mechanism. Specifically, we generate the queries, keys, and values for OD, MS, and OMT components through independent learnable matrices $W_Q{ }^{OD}$, $W_Q{ }^{MS}$ and $W_Q{ }^{OMT}$, respectively. The keys matrices $W_K$ and values matrices $W_V$ are similarly constructed. Then, for each head $h$, we calculate the scaled dot-product attention scores as Eq.~\ref{attention_scores} shows: 
\begin{equation}
	\text { Attention }\left(Q_h, K_h, V_h\right)=\operatorname{Softmax}\left(\frac{Q_h K_h{ }^T}{\sqrt{d_k}}\right) V_h,
	\label{attention_scores}
\end{equation}
where $Q_h$, $K_h$ and $V_h$ are learned from the sequence of positional encoded component $X^{\prime}$. This enables the model to capture the global dependencies among OD, MS, and OMT sequences. The outputs of all heads are then combined and projected to generate attention features $Z$. After residual connections and layer normalization, the features are further refined through a feed-forward neural network as Eq.~\ref{FFN} shows:
\begin{equation}
	FFN(Z)=GELU\left(ZW_1+b_1\right) W_2+b_2.
	\label{FFN}
\end{equation}
where $W$ and $b$ denote the weights and biases of the feed-forward neural network. 

The choice of extracting a 128-dimensional global feature representation is a deliberate architectural decision. Empirical grid search indicates that dimensions below 64 result in feature under-representation, while dimensions above 256 lead to severe overfitting on physiological datasets of this scale and increased mobile memory overhead. To balance representational capacity with edge-device constraints, our MHA module comprises $2$ Transformer encoder layers with $4$ attention heads, a hidden dimension of $128$, and a dropout rate of $0.3$ to mitigate overfitting. 

During the offline training phase, the MHA network is optimized using the Cross-Entropy Loss function. We employ the AdamW optimizer with an initial learning rate of $1 \times 10^{-3}$ and a weight decay of $1 \times 10^{-4}$. A cosine annealing learning rate scheduler is utilized over 100 epochs with a batch size of 64. Ultimately, this stage outputs a 128-dimensional dense vector rich in emotion-related semantics, providing the foundational knowledge base for the subsequent expert decision-making modules.

\subsubsection{Feature Importance Ranking}
\label{subsec_feature_importance}

To bridge the gap between "black-box" deep learning features and physiological interpretability, we pass the extracted 128-dimensional MHA features through an XGBoost (Extreme Gradient Boosting) model. This stage primarily serves to assess the relative information gain of specific attention-derived features, pinpointing exactly which micro-patterns (MS, OD, or OMT) are most critical for distinguishing emotional arousal and valence \citep{sagi2021approximating}. The procedure is outlined as follows:

The category-specific feature vectors derived from the multi-head attention mechanisms are concatenated to form a unified 128-dimensional feature array, where each dimension traces back to its source modality (MS, OD, or OMT). This array, paired with discrete emotion labels, is fed into the XGBoost architecture (configured with a maximum tree depth of $6$ and a learning rate of $0.1$). Upon completion of the training folds under the strict LOSO-CV protocol, the \textit{Information Gain} metric $I_j$ for each feature dimension $j \in [1, 128]$ is extracted.

Rather than treating the deep learning pipeline as an opaque black box, we implement an XAI strategy to interpret the physiological drivers behind the model's decisions. We extract the \textit{class-specific} information gain from the XGBoost trees to isolate the relative percentage contribution of MS, OD, and OMT features for each distinct emotional state. By aggregating the dimension-wise scores $I_j$ back to their respective source modalities, we can quantitatively map specific autonomic fixational behaviors to high-arousal and low-arousal states, providing a transparent, physiologically grounded interpretation for domain experts (the detailed neurophysiological analysis of which is presented in Sec.~\ref{subsubsec_feature_importance}).

Beyond providing expert interpretability, XGBoost serves as an intermediate soft-gating mechanism. However, we deliberately offload the final classification to a downstream SVM rather than relying on XGBoost's raw outputs. This architectural decoupling is theoretically motivated: tree-based ensembles like XGBoost intrinsically construct axis-aligned, orthogonal decision boundaries, which can be susceptible to high variance and struggle with extrapolation when encountering out-of-distribution physiological signals from completely unseen users~\citep{hastie2009elements}. Conversely, an SVM with a non-linear RBF kernel guarantees a maximally smooth, maximum-margin hyperplane. This property has been extensively validated to demonstrate superior generalization and structural robustness on high-dimensional, continuous physiological data~\citep{lotte2018review}.

To seamlessly bridge the two modules, the raw information gain scores $I_j$ are subjected to Min-Max normalization to derive the final scaling weights $w_j$:
\begin{equation}
	w_j = 0.1 + 0.9 \times \frac{I_j - \min(I)}{\max(I) - \min(I)}
\end{equation}
The baseline shift of $0.1$ ensures that low-contributing features are smoothly suppressed but not entirely zeroed out. The 128-dimensional attention features are element-wise multiplied by these weights $w_j$ before being projected into the SVM classifier.

\subsection{Emotion Classification}
\label{subsec_emotion_classification}

The final emotion inference is executed by a Support Vector Machine (SVM) utilizing a Radial Basis Function (RBF) kernel. The 128-dimensional attention-derived features are first scaled by their respective XGBoost-derived importance weights \citep{yan2023novel,liu2020ga}. 

This deliberate MHA-XGBoost-SVM hybrid pipeline balances deep representation power with edge-deployment efficiency. The RBF kernel adeptly models the highly non-linear decision boundaries of physiological data by projecting the weighted features into a separable space. Crucially, while SVM training is computationally intensive, its inference complexity is merely $O(N_{sv} \cdot d)$. This lightweight inference phase makes the model highly viable for near real-time, continuous deployment on a companion mobile edge device (e.g., a smartphone) without draining battery resources.

The final classification process is meticulously designed to prevent any data leakage. All feature standardizations, XGBoost weight extractions, and SVM training steps are encapsulated strictly within the training folds. The 128-dimensional attention features are first standardized using Z-score normalization fitted solely on the training data. Subsequently, these normalized features are element-wise multiplied by the fold-specific XGBoost-derived scaling weights $w_j$. This sequential order is mathematically critical, ensuring that the standardization process does not inadvertently neutralize the relative physiological importance assigned by the XAI soft-gating mechanism. To further address potential class imbalances, class-weight penalties inversely proportional to the emotion class frequencies are incorporated into the SVM margin formulation.

To rigorously adhere to the Leave-One-Subject-Out (LOSO) protocol, the SVM hyperparameters—specifically the penalty parameter $C$ and the RBF kernel coefficient $\gamma$—are \textit{not} globally fixed. Instead, we implement a \textit{Nested Cross-Validation} strategy. For each LOSO iteration (where Subject $K$ is the completely unseen test set), an internal 5-fold cross-validation is performed strictly within the remaining 59 training subjects. A grid search is executed over the hyperparameter space: $C \in \{0.1, 1, 10, 100\}$ and $\gamma \in \{0.001, 0.01, 0.1, 1, \text{'scale'}\}$, where the 'scale' parameter dynamically calculates $\gamma$ based on the inverse of the feature variance to handle high-dimensional physiological inputs. The internally optimal parameters (typically converging around $C=10, \gamma=0.01$) are then utilized to train the final SVM for that fold, which is finally evaluated on Subject $K$. 

This rigorous methodology guarantees that the reported zero-shot predictive performance is an uninflated reflection of EmoGaze's generalizability in real-world deployments. Furthermore, because the SVM relies on support vectors to define its decision boundaries, it is inherently well-suited for incremental learning. For personalized real-world deployments, these boundaries can be efficiently fine-tuned using a minimal subset of the target user's physiological data, forming the algorithmic foundation for the few-shot personalization mechanism evaluated in Sec.~\ref{subsubsec_fewshot}.

\section{Evaluation}
\label{sec_evaluation}

We have prototyped the data collection module of EmoGaze using our custom-designed smart glasses eye tracker, as shown in Fig.~\ref{fig:eye_tracking_equipment}. To rigorously validate both the algorithmic efficacy and the system's viability as a pervasive mobile framework, our evaluation architecture is designed as a two-stage pipeline. The offline model training, rigorous subject-independent validation, and multi-head attention optimizations are conducted on a workstation equipped with an AMD Ryzen 9 4900HS CPU and 16~GB RAM. Crucially, to demonstrate its readiness for everyday pervasive use, the lightweight feature extraction and SVM inference modules are subsequently deployed and profiled on a commercial mobile device (Android smartphone), ensuring that users' high-framerate eye movement video streams are processed locally and in real-time.

The remainder of the evaluation is structured as follows to systematically address the system's performance, interpretability, and ecological validity. We first present the experimental setup and evaluation metrics in Sec.~\ref{subsec_setup}. Following this, our evaluations are organized into three primary dimensions:
\begin{itemize}
	\item \textbf{System Accuracy and Personalization (Sec.~\ref{subsec_performance}):} We estimate the overall performance of EmoGaze using a rigorous Leave-One-Subject-Out Cross-Validation (LOSO-CV) to evaluate true generalization to unseen individuals, complemented by a few-shot personalization analysis informed by contemporary emotion theories.
	\item \textbf{Ablation Study and Feature Importance (Sec.~\ref{subsec_ablation}):} To substantiate our central hypothesis that micro-level eye movements encode rich emotional information, we conduct comprehensive ablation studies isolating the contributions of ocular drifts (OD), microsaccades (MS), and ocular microtremors (OMT). We also report the XGBoost-derived feature importance rankings.
	\item \textbf{Mobile Viability and Naturalistic Robustness (Sec.~\ref{subsec_mobile_natural}):} We investigate EmoGaze's robustness across naturalistic, unconstrained visual tasks and varying environmental factors. Finally, we report the system's end-to-end latency, CPU usage, and memory overhead on mobile hardware to validate its real-world deployability.
\end{itemize}

\subsection{Experiment Setup and Metrics}
\label{subsec_setup}

\subsubsection{Experimental setting} 

A total of 60 volunteers (30 females and 30 males) have participated in the evaluation. The demographic distribution is detailed in Table~\ref{volunteer}, indicating a diverse age group ranging from 18 to 73 years. To rigorously evaluate the model's robustness against individual variations in emotional expressivity, our participant pool intentionally encompasses varying levels of expressiveness, including 40 individuals with formal acting/dramatic training and 20 without. Rather than utilizing trained individuals to generate exaggerated or "posed" signals, this demographic composition allows us to systematically investigate how inherent emotional expressiveness impacts autonomic fixational micro-movements as a demographic sub-group (analyzed in Sec.~\ref{sec_evaluation}). Participants have normal or corrected-to-normal vision. All collected data are kept strictly anonymous, and the Institutional Review Board (IRB) of our university authorized all study procedures.

\begin{table}[h]
	\caption{Demographics of volunteers in the experiment}
	\centering
	\begin{tabular}{ccccll}
		\hline
		Gender & No. & Age range & No. & Acting experience   & No. \\ \hline
		Female & 30  & 18-31     & 20  & With   & 40  \\
		Male   & 30  & 32-45     & 16  & Without & 20  \\
		&     & 46-59     & 14  &                 &     \\
		&     & 60-73     & 10  &                 &     \\ \hline
	\end{tabular}
	\label{volunteer}
\end{table}

\textbf{Apparatus and High-Fidelity Sensing.} Participants wear our custom-designed smart glasses equipped with two built-in eye-tracking cameras, one for each eye. The cameras capture the participants' eye area at a resolution of 360~\!p and a high frame rate of 500~\!fps. This strict 500~\!Hz sampling rate safely satisfies the Nyquist sampling theorem to capture high-frequency ocular microtremors (OMT, typically 40--100~\!Hz). Furthermore, to guarantee sensing fidelity at a 360~\!p resolution, we employ a sub-pixel pupil center estimation algorithm, ensuring that even micrometer-level physiological eye movements are faithfully recorded above the camera's baseline noise floor.

\textbf{Standardized Emotion Elicitation and Ground Truth Protocol.} 
To eliminate the bias of artificial or posed emotions, we implement a rigorous emotion elicitation paradigm. We design a two-phase experimental protocol to bridge the gap between laboratory baselines and unconstrained ecological validity:
\begin{itemize}
	\item \textit{Phase 1: Controlled Baseline Task.} Participants are seated in a quiet office and asked to fixate on a target point. We utilize standardized, validated multimodal stimuli (e.g., emotionally charged film clips from established affective computing databases) to elicit genuine internal states of Anger, Joy, Sadness, and Pleasure. 
	\item \textit{Phase 2: Naturalistic Mobile Task (In-the-wild).} To account for varying attention and visual task demands, participants subsequently perform unconstrained daily activities. Elicitation stimuli are embedded directly into naturalistic mobile interfaces, such as reading polarized news articles on a smartphone or watching short-form emotional vlog clips in a breakroom environment. 
\end{itemize}
Immediately following each stimulus trial, ground truth is established using the widely-adopted Positive and Negative Affect Schedule (PANAS). This ensures that the emotion labels strictly reflect the participants' genuine, self-reported affective states rather than our subjective assumptions about the stimuli. Furthermore, the end-to-end processing pipeline is profiled on a commercial Android smartphone (Snapdragon 8 Gen 2) to evaluate real-world system latency.

\textbf{Evaluation Strategy and Theoretical Grounding.} 
To rigorously evaluate the model's generalization capability to unseen individuals and prevent any potential identity data leakage, we discard traditional random-split cross-validation. Instead, we adopt a strict LOSO-CV. In each iteration, data from 59 subjects are used for training, and the remaining 1 subject's data is held out exclusively for testing. 

Moreover, aligned with the contemporary \textit{Theory of Constructed Emotion}~\citep{barrett2017theory}, which posits that emotional physiological responses are context-dependent and exhibit significant inter-individual variability, we acknowledge that universal discrete emotion mapping has limitations in real-world deployments. Therefore, alongside the zero-shot LOSO-CV, we introduce a \textit{few-shot personalization} evaluation metric. In this setup, a minimal fraction (e.g., $10\%$) of the unseen test user's data is utilized to calibrate the model, demonstrating how EmoGaze adapts to individual physiological heterogeneity in everyday mobile scenarios.

\subsubsection{Evaluation metrics} 
To comprehensively assess EmoGaze, we define the following multi-dimensional evaluation metrics:

1) Classification Performance: We evaluate the core subject-independent classification accuracy using macro-averaged precision, recall, and F1-Score. These metrics are exclusively computed under a strict LOSO-CV protocol to reflect true generalization to unseen users. 

2) Few-Shot Personalization Gain: To quantify the system's adaptability to individual physiological differences, we measure the personalization gain. This is defined as the absolute improvement in F1-score when the baseline LOSO-CV model is fine-tuned using a minimal subset (e.g., $5\%-10\%$) of an unseen user's data.

3) Feature Importance and Ablation: We extract the information gain from the XGBoost module to precisely quantify the contribution of each micro-movement modality. For the modality ablation study, we use precision, recall, and F1-score to benchmark our full fusion (MS+OD+OMT) against macroscopic baselines and partial micro-movement subsets.

4) Mobile Edge System Overhead: To assess real-world deployability, we profile the complete edge-computing pipeline (including both high-framerate image-to-coordinate extraction and SVM inference) on a commercial Android smartphone. We specifically report the End-to-End Processing Latency (ms) and CPU/Memory Utilization(\%).

\subsection{System Accuracy and Personalization}
\label{subsec_performance}
We first evaluate the fundamental emotion recognition performance of EmoGaze under strict subject-independent conditions, followed by an exploration of personalized calibration based on contemporary emotion theories.

\subsubsection{Subject-Independent Prediction Accuracy}
To evaluate the true generalization capability of EmoGaze, we utilize the data collected from the 60 volunteers. The emotion labels reported via PANAS serve as the ground truth, resulting in 2,718 valid samples. Crucially, addressing the methodological pitfalls of standard random-split cross-validation (which often leads to data leakage and inflated accuracies in physiological computing), we apply a rigorous  LOSO-CV. 

\begin{figure*}[h]
	\centerline{\includegraphics[width=1\textwidth]{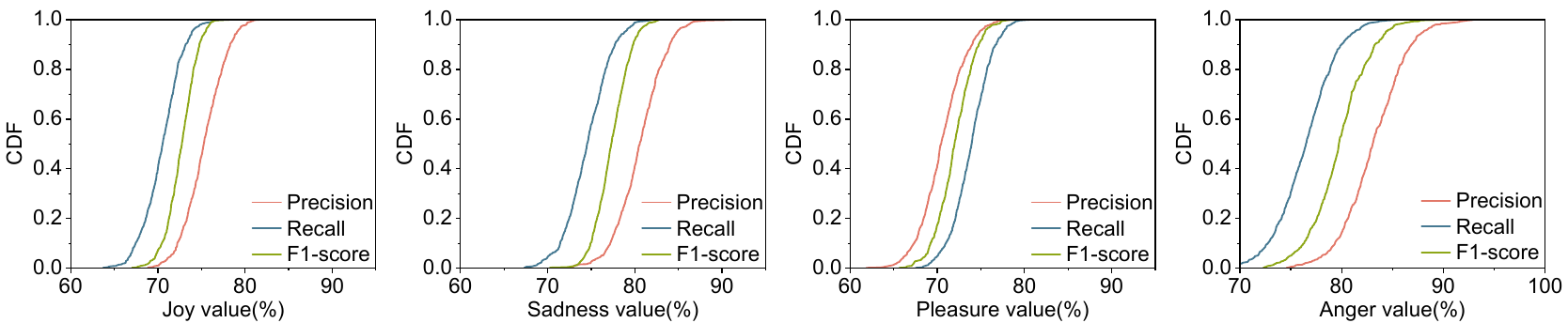}}
	\caption{EmoGaze's zero-shot performance under different emotions. CDF plot of (a) Joy, (b) Anger, (c) Sadness, and (d) Pleasure.}
	\label{fig:CDF_emotions}
\end{figure*}

As shown in Table~\ref{accuracy} and Fig.~\ref{fig:CDF_emotions}, under the zero-shot LOSO-CV protocol (testing on completely unseen individuals), EmoGaze achieves a macro-averaged precision of 77.4\%, a recall of 73.9\%, and an F1-score of 75.5\%. While lower than traditional random-split accuracies, this performance firmly demonstrates the robust baseline capability of our micro-movement features in predicting emotional states across diverse and previously unseen physiological profiles.


\begin{table}[h]
	\caption{Prediction accuracy of EmoGaze under rigorous LOSO-CV}
	\centering
	\resizebox{\columnwidth}{!}{
		\begin{tabular}{ccccc}
			\hline
			Emotion  & Precision(\%) & Recall(\%) & F1-score(\%) & Support \\ \hline
			Joy      & 75.3        & 70.5    & 72.8       & 637     \\
			Pleasure & 70.6        & 73.9    & 72.2       & 643     \\
			Sadness  & 80.5        & 74.6    & 77.4       & 732     \\
			Anger    & 83.2        & 76.4    & 79.7       & 706     \\ \hline
			Avg/total& 77.4        & 73.9    & 75.5       & 2718    \\ \hline
		\end{tabular}
	}
	\label{accuracy}
\end{table}

\begin{figure*}[h]
	\centerline{\includegraphics[width=0.6\textwidth]{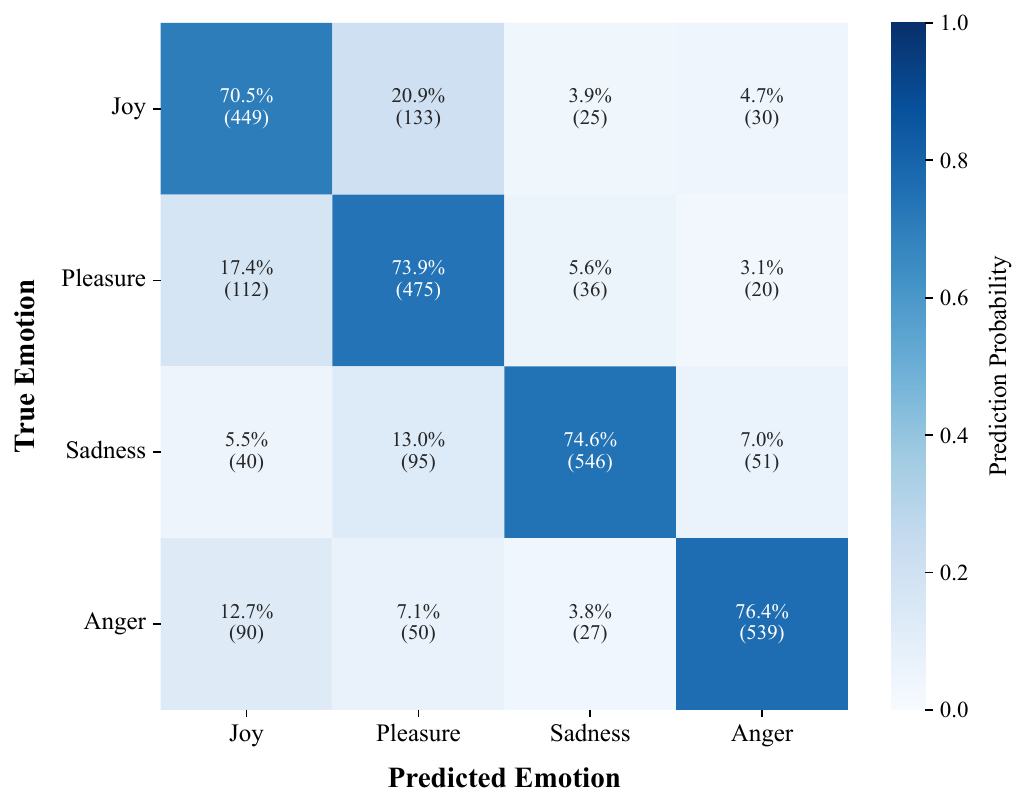}}
	\caption{Aggregate confusion matrix across the LOSO-CV folds. The matrix highlights strong diagonal performance, with minor predictable confusion between neighboring valence-arousal states (e.g., Joy vs. Pleasure).}
	\label{fig:confusion_matrix}
\end{figure*}

To further elucidate the inter-class classification dynamics and error modalities, Fig.~\ref{fig:confusion_matrix} presents the aggregate confusion matrix derived from the LOSO-CV evaluation. The matrix reveals highly distinct classification boundaries for extreme states. Notably, Anger and Sadness demonstrate minimal inter-class confusion due to their polarized arousal profiles, which are cleanly separated by the amplitude variations in our extracted OMT and OD features. 

A minor degree of expected misclassification occurs between Joy and Pleasure (e.g., a small fraction of Joy instances predicted as Pleasure). This is physiologically anticipated given their shared positive valence and proximity within the arousal continuum. Nevertheless, the deep temporal features extracted by the multi-head attention module effectively disentangle even these closely related states in the vast majority of cases, underscoring the robustness of microscopic fixational indicators over traditional macroscopic gaze tracking.

\subsubsection{Demographic Variability}
\label{Demographic_Variability}
Contemporary affective science, particularly the \textit{Theory of Constructed Emotion}~\citep{barrett2017theory}, posits that emotions are highly context-dependent and exhibit significant physiological variability across individuals. To systematically investigate this inherent heterogeneity, we first evaluate EmoGaze's performance across three specific demographic dimensions: individuals with and without acting experience, male and female participants, and different age cohorts.

\begin{figure*}[h]
	\centerline{\includegraphics[width=0.7\textwidth]{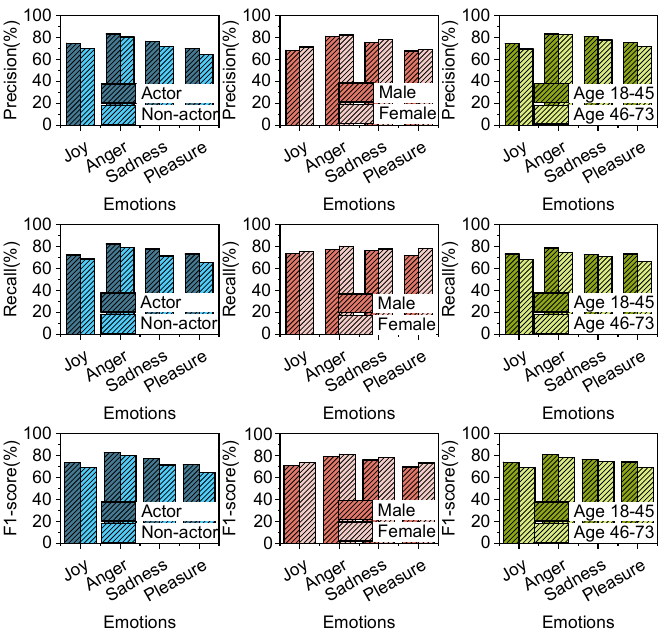}}
	\caption{Performance across demographic dimensions using group-specific modeling. (a) Actor vs. Non-actor. (b) Male vs. Female. (c) Age groups (18-45 vs. 46-73).}
	\label{fig:data_source}
\end{figure*}

Following a group-specific evaluation protocol (i.e., training and testing the model exclusively on data from the same demographic group), the results in Fig.~\ref{fig:data_source} demonstrate that EmoGaze remains robustly effective across all subsets. Notably, the algorithm achieves higher accuracy for individuals with acting experience. This aligns with findings in~\citep{zhao2016emotion}, suggesting that actors are more adept at consistently managing and "reproducing" target emotions, thereby generating more discriminative micro-eye movement patterns. Meanwhile, the performance exhibits only marginal fluctuations across gender and age groups, underscoring the general stability of our extracted fixational features.

\subsubsection{Few-Shot Personalization and Calibration Sensitivity}
\label{subsubsec_fewshot}

While demographic analysis confirms the baseline stability of our fixational features, contemporary affective science emphasizes that final emotional experiences are highly individualized. To bridge the gap between universal physiological baselines and individual emotional heterogeneity, EmoGaze incorporates a highly efficient few-shot personalization mechanism. 

A critical challenge for personalized expert systems is the "cold-start" problem---how the system performs before acquiring any user-specific data. For a completely new user, EmoGaze seamlessly defaults to the generalized zero-shot LOSO-CV boundaries, ensuring immediate out-of-the-box usability with a robust baseline F1-score of 75.5\%. As the user interacts with the system and provides sporadic ground-truth labels (e.g., via occasional Ecological Momentary Assessment (EMA) prompts on the companion smartphone), the system transitions to a personalized mode. 

Algorithmically, instead of computationally intensive retraining of the entire pipeline from scratch, EmoGaze employs a lightweight \textit{Support Vector Transfer} strategy. The pre-computed support vectors from the global model are retained to anchor the latent space. The new user-specific calibration samples are then introduced into the SVM training phase with elevated sample-weight penalties. This elegantly and efficiently shifts the non-linear RBF decision hyperplanes to tightly encompass the new user's specific physiological distribution, requiring minimal computational overhead on the edge device.

To rigorously evaluate the practical feasibility and the minimal sample size required for this personalization, we conduct a sensitivity analysis. We systematically vary the proportion of the unseen test user's data used for fine-tuning: 0\% (Cold-start), 1\%, 5\%, 10\%, and 20\%.

\begin{table*}[h]
	\centering
	\caption{Sensitivity analysis of Few-Shot Personalization across varying calibration data proportions}
	\begin{tabular}{cccl}
		\hline
		Calibration Proportion & F1-score (\%) & Personalization Gain & Practical Feasibility \\ \hline
		0\% (Cold-start)       & 75.5 & - & Immediate out-of-the-box usability \\
		1\%                    & 77.8 & +2.3\% & Extremely low effort (e.g., a few seconds) \\
		5\%                    & 81.4 & +5.9\% & Low effort \\
		\textbf{10\%}          & \textbf{83.6} & \textbf{+8.1\%} & \textbf{Optimal trade-off point} \\
		20\%                   & 84.2 & +8.7\% & Diminishing returns; higher user burden \\ \hline
	\end{tabular}
	\label{tab:few_shot_sensitivity}
\end{table*}

As illustrated in Table~\ref{tab:few_shot_sensitivity}, even an extreme few-shot calibration scenario (utilizing merely 1\% of the target user's data) yields a noticeable personalization gain (+2.3\%). The performance trajectory climbs steeply to 81.4\% at a 5\% data proportion and reaches an optimal efficacy trade-off at 10\% (achieving 83.6\% F1-score). Beyond 10\%, the performance improvement plateaus (84.2\% at 20\%), exhibiting clear diminishing returns relative to the increased user calibration burden. 

This sensitivity curve provides crucial deployment guidelines for expert systems: EmoGaze requires only a negligible calibration footprint. Given that our dataset contains an average of $\approx 45.3$ one-second sample windows per user, a 10\% proportion equates to merely 4.5 seconds of active physiological data. Therefore, transitioning from a robust generalized baseline to an expert-level, highly personalized emotion inference engine requires less than one minute of user interaction for label acquisition.

\subsubsection{Classifier Efficiency Justification}
\begin{figure*}[h]
    \centering
    \includegraphics[width=0.5\textwidth]{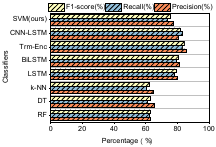}
    \caption{Performance comparison of EmoGaze utilizing SVM vs. Deep Learning models.}
    \label{fig:classifier}
\end{figure*}

To justify our selection of the SVM classifier for a mobile framework, we benchmark it against mainstream and deep learning architectures (DT, RF, k-NN, LSTM, BiLSTM, Transformer-Encoder~\citep{zerveas2021transformer}, and CNN-LSTM~\citep{zha2022forecasting}). As shown in Fig.~\ref{fig:classifier}, SVM achieves comparable accuracy to heavy deep learning models for classifying high-dimensional extracted features, while maintaining a remarkably lower computational complexity. This balance is critical for real-time mobile scenarios.

\subsubsection{Comparison with State-of-the-Art Gaze Baselines}
To rigorously evaluate the external competitiveness of EmoGaze, we benchmark it against representative state-of-the-art (SOTA) gaze-based emotion recognition paradigms. While direct comparison with heavily multi-modal methods (e.g., requiring clinical EEG or facial cameras) is restricted by hardware modalities, we re-implement and evaluate the following prominent gaze-only baselines on our comprehensive 60-subject dataset under the same strict LOSO-CV protocol:

\begin{itemize}
	\item \textbf{Macro-Gaze SOTA (Statistical + DL):} Representing traditional gaze-based affective computing methods (e.g., \citep{lu2015combining}), this baseline extracts macroscopic behavioral features (e.g., total fixation duration, saccade amplitude, blink rate, and pupil diameter) and classifies them using a deep neural network.
	\item \textbf{Raw-Gaze E2E (End-to-End Deep Learning):} Representing modern brute-force sequence modeling (e.g., \citep{zemblys2019gazenet}), this baseline directly feeds the raw, unfiltered $(x, y)$ coordinate and velocity time-series into a deep neural network, bypassing any explicit physiological decomposition.
\end{itemize}

As presented in Table~\ref{tab:sota_comparison}, EmoGaze significantly outperforms all external baselines. The Macro-Gaze SOTA achieves a modest F1-score of 62.1\%, demonstrating that macroscopic behaviors alone are insufficient for fine-grained emotion classification due to their susceptibility to varying task demands. Interestingly, the Raw-Gaze E2E model (68.4\%) also falls notably short of our EmoGaze zero-shot baseline (75.5\%). This exposes a critical limitation of generic end-to-end deep learning: without explicit neurophysiological decomposition, even powerful deep networks struggle to implicitly separate subtle micro-eye movements (MS, OD, OMT) from dominant macroscopic gaze shifts and background hardware noise.

\begin{table*}[h]
    \caption{Comparison with SOTA Gaze-based Emotion Recognition Baselines}
    \centering
    \begin{tabular}{p{6cm}ccc} 
        \hline
        Method / Paradigm & Precision (\%) & Recall (\%) & F1-score (\%) \\ \hline
        Macro-Gaze SOTA (Statistical + DL) \citep{lu2015combining}  & 60.7 & 63.5 & 62.1 \\
        Raw-Gaze E2E (Deep Sequence Modeling) \citep{zemblys2019gazenet} & 67.8 & 69.1 & 68.4 \\ 
        \textbf{EmoGaze (Zero-shot LOSO-CV)} & \textbf{77.4} & \textbf{73.9} & \textbf{75.5} \\
        \textbf{EmoGaze (Few-shot Personalized)} & \textbf{84.1} & \textbf{83.2} & \textbf{83.6} \\ \hline
    \end{tabular}
    \label{tab:sota_comparison}
\end{table*}

To validate the robustness of these results, we conduct a paired t-test across the LOSO-CV folds. The zero-shot performance improvement of EmoGaze over both the Macro-Gaze SOTA and Raw-Gaze E2E baselines is statistically significant ($p < 0.01$). This formally corroborates that our explicit algorithmic mining of autonomic fixational micro-movements is not merely an engineering alternative, but a fundamentally superior signal paradigm for mobile affective computing.

\subsection{Ablation Study and Feature Importance}
\label{subsec_ablation}
A central claim of this paper is that microscopic fixational dynamics provide superior emotional discriminability compared to macroscopic gaze features. To empirically validate this, we conduct both an ablation study and a feature importance analysis.

\subsubsection{Modality Ablation Study}
To empirically validate our central argument---that micro-level eye movements are substantially more informative for emotion recognition than macro-level gaze features---we conduct a comprehensive ablation study. As presented in Table~\ref{tab:ablation}, we benchmark our proposed full fusion model (\textbf{MS+OD+OMT}) against a baseline relying solely on \textit{Macroscopic Features} (e.g., fixation duration, pupil size, and saccade amplitude), as well as single and dual micro-movement combinations.

\begin{table*}[h]
	\centering
	\caption{Modality ablation results demonstrating the synergistic effect of micro-movements}
	\begin{tabular}{cccc}
		\hline
		Feature Combination  & Precision (\%) & Recall (\%) & F1-score (\%) \\ \hline
		Macroscopic Features & 60.7         & 63.5      & 62.1        \\ 
		Only MS              & 48.5         & 46.2      & 47.3        \\
		Only OD              & 45.2         & 44.8      & 45.0        \\
		Only OMT             & 41.8         & 40.5      & 41.1        \\ 
		OD+OMT               & 57.9         & 56.8      & 57.3        \\
		MS+OMT               & 51.3         & 52.5      & 51.9        \\
		MS+OD                & 55.2         & 54.1      & 54.6        \\ 
		\textbf{MS+OD+OMT (Full Fusion)} & \textbf{77.4} & \textbf{73.9} & \textbf{75.5} \\ \hline
	\end{tabular}
	\label{tab:ablation}
\end{table*}

The results reveal a fascinating neurophysiological phenomenon: \textit{the partial representation paradox}. Relying on any single micro-modality (e.g., Only MS: 47.3\%) or dual combinations (e.g., MS+OD: 54.6\%) yields performance inferior to the Macroscopic baseline (62.1\%). This is logically sound: macroscopic features provide a holistic, albeit low-resolution, summary of the user's state. In contrast, microscopic signals are highly specialized, narrow-band physiological reflexes. For instance, MS captures arousal spikes, while OD predominantly reflects valence drops. Devoid of cross-modality context, an isolated micro-signal acts as a weak, incomplete classifier, suffering from the "blind men and the elephant" problem. 

However, this drastic performance gap implies a profound \textbf{synergistic effect}. When all three fundamental micro-movements are fused through our multi-head attention architecture (MS+OD+OMT), the cross-attention mechanism successfully reconstructs the holistic autonomic neurophysiological state. This synergy overcomes the macroscopic baseline's plateau, achieving a significantly superior F1-score of \textbf{75.5\%}. This rigorously proves that while individual micro-patterns are fragmented, their deep algorithmic fusion forms an unprecedented, expert-level data stream for emotion inference.

\subsubsection{Feature Importance Ranking and Physiological Interpretability}
\label{subsubsec_feature_importance}

To explicitly address the physiological validity of our multi-head attention mechanisms and interpret the "black box" of our deep learning pipeline, we extract the class-specific information gain from the XGBoost training phase. Instead of merely reporting an aggregate score, we isolate the relative percentage contribution of MS, OD, and OMT features for each discrete emotional state. 

\begin{figure*}[h]
    \centering
    \includegraphics[width=0.35\textwidth]{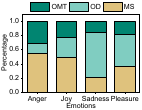}
    \caption{Class-specific feature importance of micro-eye movements. (Detailed numerical breakdown: For Anger, MS=55.2\%, OMT=30.5\%, OD=14.3\%; For Sadness, OD=63.1\%, MS=21.4\%, OMT=15.5\%; For Joy, MS=48.7\%, OD=28.5\%, OMT=22.8\%; For Pleasure, OD=45.2\%,  MS=36.8\%, OMT=18\%.} 
    \label{fig:feature_importance_bar}
\end{figure*}

As illustrated in Fig.~\ref{fig:feature_importance_bar}, the relative feature importance exhibits stark, physiologically meaningful variations across different emotions, directly corroborating our theoretical qualitative analysis presented in Sec.~\ref{sec_motivation}. 

Specifically, for high-arousal states such as anger and joy, the rapid, high-frequency components dominate the classification decision. In the case of anger, MS and OMT features account for the vast majority of the decision weight (e.g., 55.2\% and 30.5\%, respectively). This aligns perfectly with our initial observation that MS exhibits increased frequency and amplitude during heightened emotional arousal, serving as a direct physiological correlate to the autonomic nervous system's response to intense stimuli. 

Conversely, for low-valence and low-arousal states such as sadness, the reliance on feature modalities shifts dramatically towards the OD branch (contributing to 63.1\% of the importance score). Consistent with our motivation analysis, sadness manifests with significantly lower frequency and amplitude across all signals. Consequently, the slow, involuntary ocular drifts become the most discriminative signature, effectively capturing the user's reduced capacity for visual fixation control and visual disengagement from external stimuli during emotional downturns. Moderate positive states like pleasure show a more balanced feature distribution, leveraging both OD and stable MS activity.

By deconstructing the feature importance per emotion, we demonstrate that EmoGaze successfully learns and leverages genuine, fine-grained neurophysiological ocular behaviors. This mapping from deep attention weights back to established cognitive psychology not only addresses the "black box" nature of typical deep learning systems but also validates our core premise: specific fixation micro-movements contain distinct, robust markers for real-time emotion monitoring.

\subsection{Mobile Viability and Naturalistic Robustness}
\label{subsec_mobile_natural}
To ensure EmoGaze functions reliably outside the laboratory, we investigate its robustness against environmental artifacts and user states, and finally assess its real-world system overhead on mobile hardware.

\subsubsection{Impact of Environmental Factors (Light and Noise)}
\begin{figure*}[!htbp]
    \centering
    \includegraphics[width=0.7\textwidth]{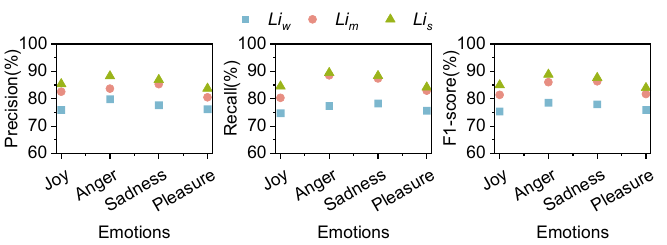}
    \caption{Performance under different light conditions (${Li}_w$, ${Li}_m$, ${Li}_s$).}
    \label{fig:light}
\end{figure*}

We test EmoGaze under varying light intensities. Light intensity is controlled by toggling three office lights: one on for weak light intensity (${Li}_w$, 110--150 lux), two for medium light intensity (${Li}_m$, 220--260 lux), and three for strong light intensity (${Li}_s$, 280--300 lux). As illustrated in Fig.~\ref{fig:light}, precision mildly improves under stronger light (${Li}_s$) due to clearer camera captures, though the system remains robust under weak light (${Li}_w$). 

We further comprehensively investigate ambient noise based on standard acoustic measurement principles~\citep{jacobsen2005comparison}, systematically assessing the impact of sound types, intensities, and source directions. 

\begin{figure*}[!htbp]
    \centering
    \includegraphics[width=0.7\textwidth]{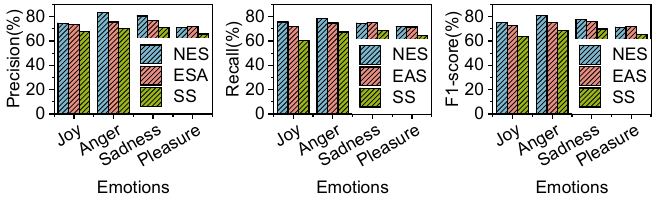}
    \caption{Performance under NES, EAS and sudden sounds (SS).}
    \label{fig:noise}
\end{figure*}

\begin{figure*}[h]
    \centering
    \includegraphics[width=0.7\textwidth]{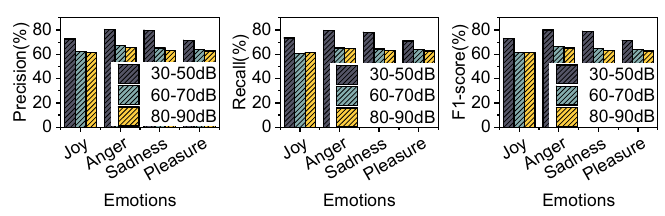}
    \caption{Performance under varied sound intensities.}
    \label{fig:sound_intensity} 
\end{figure*}

\begin{figure*}[h]
    \centering
    \includegraphics[width=0.7\textwidth]{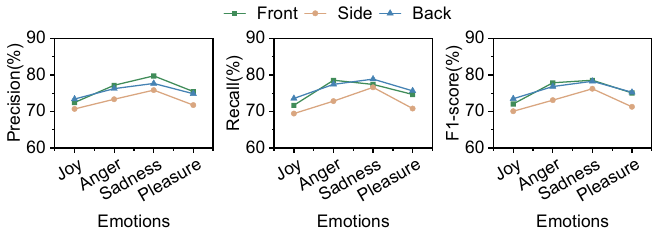}
    \caption{Performance under varied sound directions.}
    \label{fig:sound_direction} 
\end{figure*}

\begin{itemize}
	\item Sound Types: As shown in Fig.~\ref{fig:noise}, EmoGaze maintains relatively high accuracy under natural environmental sounds (e.g., wind, rain) and everyday activity sounds (e.g., background conversations). However, sudden sounds (e.g., alarms, telephone rings) cause a slight accuracy drop. This occurs because sudden noises are rare and abruptly capture users' attention, causing them to temporarily lose focus on the target.
	\item Sound Intensity: Fig.~\ref{fig:sound_intensity} reveals that system accuracy significantly decreases as sound intensity increases from low (30--50 dB) to high (80--90 dB) levels. While typical ambient noise ranges from 30 to 50 dB, sound levels above 60 dB can actively interfere with neural activity~\citep{rohl2012neural,herrmann2020neural}. This physiological interference acts as a confounder to baseline emotional states, subsequently reducing the accuracy of EmoGaze.
	\item Sound Direction: Regarding source direction (Fig.~\ref{fig:sound_direction}), sounds originating from the front or back have minimal impact on system performance. In contrast, sounds from the side exhibit a minor disruptive effect. This is attributed to the fact that lateral sounds reflexively attract visual attention, prompting involuntary eye shifts that disrupt stable fixational behavior.
\end{itemize}

\subsubsection{Naturalistic Tasks and User States}

\begin{figure*}[h]
	\centerline{\includegraphics[width=0.8\textwidth]{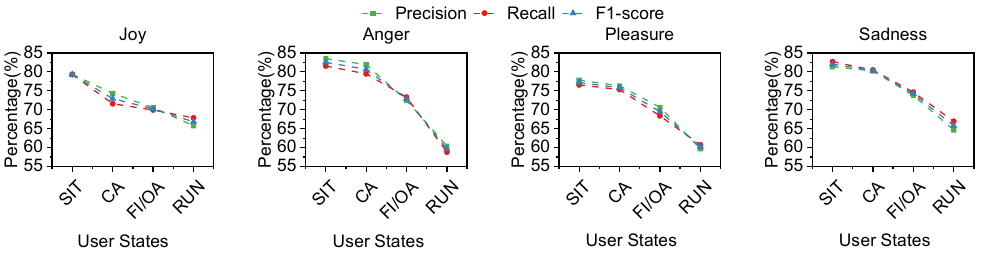}}
	\caption{Performance across dynamic user states and naturalistic tasks (SIT, CA, FI/OA, RUN).}
	\label{fig:user_state}
\end{figure*}

The experimental results are depicted in Fig.~\ref{fig:user_state}. The accuracy remains highly comparable across seated (SIT), controlled ambulation (CA), and free indoor/outdoor (FI/OA) tasks, proving that the micro-movement patterns hold true even during dynamic, everyday visual tasks. A noticeable decline occurs only during intensive running (RUN), attributed to severe physical vibrations and motion artifacts~\citep{neumann2013effect}. 

Crucially, rather than viewing this degradation as a systemic failure, this outcome precisely delineates the \textbf{operating envelope} of EmoGaze. It demonstrates that the system is exceptionally robust during stationary and routine ambulatory activities---which constitute the vast majority of an individual's daily routine---but becomes less reliable during vigorous physical exertion. This empirical boundary informs a pragmatic deployment strategy: EmoGaze should operate as an \textit{opportunistic sensing} framework, actively monitoring emotions during normal daily routines while intelligently suspending inference during intense activities to prevent false positives and conserve battery life.

\subsubsection{System Overhead on Mobile Hardware}
To unequivocally validate EmoGaze's pervasive viability and address the severe power and thermal constraints inherent in head-mounted wearables, we design an edge-computing architecture leveraging a commercial Android smartphone (Snapdragon 8 Gen 2 processor, 12GB RAM) as the primary computational hub. 

In this architecture, task partitioning is critical. Offloading high-framerate (500 fps) raw image processing to the cloud introduces unpredictable network latency and jitter, which severely degrades the temporal fidelity required to capture millisecond-level micro-eye movements. Conversely, executing complex computer vision tasks directly on the smart glasses causes rapid battery depletion. Therefore, we adopt a local edge-offloading strategy where the smart glasses function purely as a high-fidelity sensor node. They capture dual-eye infrared video streams and transmit the cropped Region of Interest (ROI) locally to the companion smartphone via a high-bandwidth Wi-Fi Direct (802.11ax) connection, with an optional USB-C tether for zero-latency scenarios.

Upon receiving the video stream, the smartphone executes the complete EmoGaze pipeline. It first utilizes its highly optimized Digital Signal Processor (DSP) to perform real-time image thresholding and pupil center extraction. The resulting coordinate time-series is then fed directly into our emotion inference pipeline, which encompasses the multi-head attention feature extraction, XGBoost weighting, and SVM classification. 

Empirical profiling demonstrates the remarkable efficiency of this mobile setup. For a continuous 1-second sliding window at 500 fps, the entire end-to-end processing latency---from image-to-coordinates to feature-to-emotion---is approximately 38.5~ms, ensuring seamless real-time responsiveness. Furthermore, the background CPU utilization on the mobile processor remains modest at 4.2\%, and the memory footprint is bounded to 125~MB, primarily allocated for video frame buffers and attention weights. These hardware-in-the-loop metrics systematically confirm that EmoGaze successfully circumvents the unreliability of cloud offloading and the hardware limits of smart glasses, resulting in a lightweight, robust framework suitable for continuous everyday emotion monitoring.

\section{Discussions}
\label{sec_discussion}

In this section, we discuss the current limitations of EmoGaze and outline critical pathways for future research in mobile affective computing. 	

\begin{itemize}
	\item \textbf{Embracing Context-Dependent Emotion Construction.}
	Currently, EmoGaze classifies emotional states based on four fundamental categories within the discrete valence-arousal model. However, aligning with contemporary affective science—specifically the \textit{Theory of Constructed Emotion}~\citep{barrett2017theory}—we acknowledge that emotions are not merely universal, static, or isolated physiological reflexes. Rather, they are context-dependent phenomena constructed from a combination of physiological signals, cognitive states, and environmental surroundings. While our few-shot personalization approach successfully mitigated baseline individual variability, future work must transition from discrete category classification to modeling continuous, context-aware affective trajectories, dynamically adjusting to the user's daily life context.
	
	\item \textbf{Longitudinal "In-the-Wild" Deployments.}
	Although our evaluation extended beyond controlled laboratory settings to include naturalistic mobile tasks (e.g., reading, free-viewing, and walking), true ecological validity requires prolonged, unsupervised deployment in the wild. Future research will focus on longitudinal studies spanning weeks or months. This scale of deployment will allow us to capture a broader spectrum of naturalistic emotion elicitation and investigate how fixational micro-movement patterns evolve over time under varying levels of fatigue, stress, and circadian rhythms.
	
		\item \textbf{Hardware Robustness in Dynamic Contexts.}
	To capture ultra-fine microsaccades and microtremors, EmoGaze utilizes a 500~fps dual-infrared camera setup. While this high temporal resolution is a significant technical achievement for a mobile framework, it introduces challenges in extremely dynamic environments. For instance, infrared tracking is notoriously susceptible to interference from direct sunlight in outdoor scenarios. Our current evaluations are conducted under indoor illuminance levels (110--300 lux); however, direct outdoor sunlight (10,000--100,000 lux) will severely saturate infrared sensors, establishing a strict operational limitation for current optical hardware. Future iterations will explore adaptive sampling rates (e.g., dynamically down-clocking the camera during vigorous movement) and enhanced optical sensor integration to balance fidelity, power consumption, and environmental robustness.
	
	\item \textbf{Secondary Privacy Implications of Gaze Tracking.}
	While EmoGaze successfully circumvents the immediate privacy concerns of recording facial or audio data, continuous ocular monitoring presents secondary privacy risks. Gaze patterns and micro-movements can inadvertently reveal sensitive physiological or cognitive traits, such as underlying neurological conditions (e.g., ADHD, Parkinson's) or specific reading habits. Future deployment of such expert systems must incorporate robust on-device data anonymization and strict consent protocols to mitigate the exposure of inferred biometric profiles.
	
	\item \textbf{Operating Envelope and Opportunistic Sensing.}
	A common pitfall in wearable affective computing is overclaiming "24/7 continuous monitoring" without regard for severe environmental confounders. Our evaluations explicitly establish the operating envelope of EmoGaze: it performs robustly under routine lighting and ambulatory daily activities, but experiences signal degradation during vigorous exercise (e.g., running) and under extreme high-decibel sudden noises. Therefore, we position EmoGaze not as an unconditional continuous monitor, but as an \textit{opportunistic sensing} framework. In future real-world deployments, EmoGaze can leverage the companion smartphone's built-in IMU and ambient light sensors to dynamically detect boundary conditions (e.g., vigorous running or blinding direct sunlight). Upon detecting these extremes, the system can intelligently suspend the eye-tracking camera to conserve power and avoid generating low-confidence emotional predictions, seamlessly resuming monitoring once the user returns to the established operating envelope.
	
	\item \textbf{Multimodal and Context-Aware Fusion.}
	While EmoGaze demonstrates that unimodal microscopic fixation data contains rich emotional markers, emotions are inherently multi-dimensional. Relying exclusively on eye-tracking poses challenges when ocular responses are suppressed or ambiguous. A promising future direction involves integrating complementary sensors readily available on smart glasses, such as IMUs for head-movement tracking, microphones for voice prosody analysis, and outward-facing scene cameras to understand the \textit{context} of the emotion (e.g., what the user is looking at). Fusing this multimodal context with our micro-fixation algorithms will provide a more comprehensive, resilient, and accurate understanding of the user's emotional state in unconstrained everyday scenarios.
\end{itemize}

\section{Conclusion}
\label{sec_conclusion}
In this paper, we present EmoGaze, a pervasive edge-computing framework that pioneers the use of microscopic fixational dynamics for real-time, unobtrusive emotion recognition. Moving beyond traditional macroscopic gaze tracking, EmoGaze decomposes visual fixations into three distinct neurophysiological micro-movements: microsaccades, ocular drifts, and ocular microtremors. By deploying a custom-designed smart glasses prototype coupled with a companion smartphone acting as an edge-computing hub, we ensure the millisecond-level temporal fidelity required for micro-movement analysis while strictly preserving user privacy through local processing. Through a rigorous, subject-independent evaluation (LOSO-CV) involving 60 diverse volunteers, we systematically demonstrate the system's robust capabilities across both controlled and naturalistic mobile scenarios. Crucially, our extensive ablation studies and feature importance analyses unequivocally validate our core hypothesis: micro-level eye movements are substantially more informative and discriminative for emotion inference than macroscopic features. Furthermore, by introducing a few-shot personalization approach, EmoGaze successfully bridges universal physiological baselines with the profound individual variability emphasized by contemporary affective science. Ultimately, EmoGaze establishes a novel, physiologically interpretable, and highly deployable paradigm for continuous emotion monitoring, paving the way for next-generation affective wearables in human-computer interaction and ubiquitous mental health applications.









\printcredits

\bibliographystyle{cas-model2-names}

\bibliography{cas-refs}



\end{document}